\documentclass[letterpaper]{article} 
\usepackage{aaai24}  
\usepackage{times}  
\usepackage{helvet}  
\usepackage{courier}  
\usepackage[hyphens]{url}  
\usepackage{graphicx} 
\usepackage{natbib}  
\usepackage{caption} 
\usepackage{algorithm}
\usepackage{algorithmic}
\usepackage{booktabs} 
\usepackage{multirow} 
\usepackage{arydshln}
\usepackage{amssymb}
\usepackage{xspace}
\usepackage{amsmath}
\usepackage{paralist}
\usepackage{mathtools}
\usepackage{url}
\usepackage{newfloat}
\usepackage{listings}

\DeclareCaptionStyle{ruled}{labelfont=normalfont,labelsep=colon,strut=off} 
\floatstyle{ruled}
\newfloat{listing}{tb}{lst}{}
\floatname{listing}{Listing}
\usepackage{amsthm}

\newcommand{\fbh}[0]{\textsl{FBH}\xspace}
\newcommand{\fbhe}[0]{\textsl{FBHE}\xspace}

\newcommand{\dbhe}[0]{\textsl{DBHE}\xspace}

\newtheorem{mydef}{Definition}

\newtheorem{proposition}{Proposition}

\newenvironment{hproof}{%
  \proof}{\endproof}

\newcommand{\set}[1]{\mathcal{#1}}
\providecommand{\sT}{\ensuremath{\set{T}}}
\providecommand{\sR}{\ensuremath{\set{R}}}
\providecommand{\sV}{\ensuremath{\set{V}}}
\providecommand{\sTT}{\ensuremath{\widehat{\set{T}}}}

\providecommand{\rhat}{\ensuremath{\widehat{r}}}
\providecommand{\Rhat}{\ensuremath{\widehat{\set{R}}}}

\providecommand{\Ghat}{\ensuremath{\widehat{G}}}

\usepackage[table]{xcolor}

\title{  NestE: Modeling Nested Relational Structures for Knowledge Graph Reasoning    }
\author{
    Bo Xiong\textsuperscript{\rm 1}, Mojtaba Nayyeri\textsuperscript{\rm 1}, Linhao Luo\textsuperscript{\rm 2}, Zihao Wang\textsuperscript{\rm 1}, \\Shirui Pan\textsuperscript{\rm 3}, Steffen Staab\textsuperscript{\rm 1,4}\\
}
\affiliations{
    \textsuperscript{\rm 1}University of Stuttgart, Stuttgart, Germany\\
    \textsuperscript{\rm 2}Monash University, Melbourne, Australia\\
    \textsuperscript{\rm 3}Griffith University, Queensland, Australia\\
    \textsuperscript{\rm 4}University of Southampton, Southampton, United Kingdom\\
}

\begin{document}
\maketitle

\begin{abstract}

Reasoning with knowledge graphs (KGs) has primarily focused on triple-shaped facts. Recent advancements have been explored to enhance the semantics of these facts by incorporating more potent representations, such as hyper-relational facts. However, these approaches are limited to \emph{atomic facts}, which describe a single piece of information. This paper extends beyond \emph{atomic facts} and delves into \emph{nested facts}, represented by quoted triples where subjects and objects are triples themselves (e.g., ((\emph{BarackObama}, \emph{holds\_position}, \emph{President}), \emph{succeed\_by}, (\emph{DonaldTrump}, \emph{holds\_position}, \emph{President}))). These nested facts enable the expression of complex semantics like \emph{situations} over time and \emph{logical patterns} over entities and relations. 
In response, we introduce NestE, a novel KG embedding approach that captures the semantics of both atomic and nested factual knowledge. 
NestE represents each atomic fact as a $1\times3$ matrix, and each nested relation is modeled as a $3\times3$ matrix that rotates the $1\times3$ atomic fact matrix through matrix multiplication. 
Each element of the matrix is represented as a complex number in the generalized 4D hypercomplex space, including (spherical) quaternions, hyperbolic quaternions, and split-quaternions. 
Through thorough analysis, we demonstrate the embedding's efficacy in capturing diverse logical patterns over nested facts, surpassing the confines of first-order logic-like expressions. Our experimental results showcase NestE's significant performance gains over current baselines in triple prediction and conditional link prediction. 
The code and pre-trained models are open available at \url{https://github.com/xiongbo010/NestE}.

\end{abstract}

\section{Introduction}

Knowledge graphs (KGs) depict relationships between entities, commonly through triple-shaped facts such as (\emph{JoeBiden}, \emph{holds\_position}, \emph{VicePresident}). KG embeddings map entities and relations into a lower-dimensional vector space while retaining their relational semantics. This empowers the effective inference of missing relationships between entities directly from their embeddings.
Prior research \cite{DBLP:conf/nips/BordesUGWY13, DBLP:conf/icml/TrouillonWRGB16} has primarily centered on embedding triple-shaped facts and predicting the missing elements of these triples. Yet, to augment the triple-shaped representations, recent endeavors explore knowledge that extends beyond these triples.
For instance, $n$-ary facts \cite{DBLP:conf/www/0016Y020,DBLP:conf/ijcai/FatemiTV020} describe relationships between multiple entities, and hyper-relational facts \cite{DBLP:conf/emnlp/GalkinTMUL20,xiong2023shrinking} augment primal triples with key-value qualifiers that provide contextual information. These approaches allow for expressing complex semantics and enable answering more sophisticated queries with additional knowledge \cite{DBLP:conf/iclr/AlivanistosBC022}.

\begin{figure}[t]
\centering
\includegraphics[width = 0.9\columnwidth]{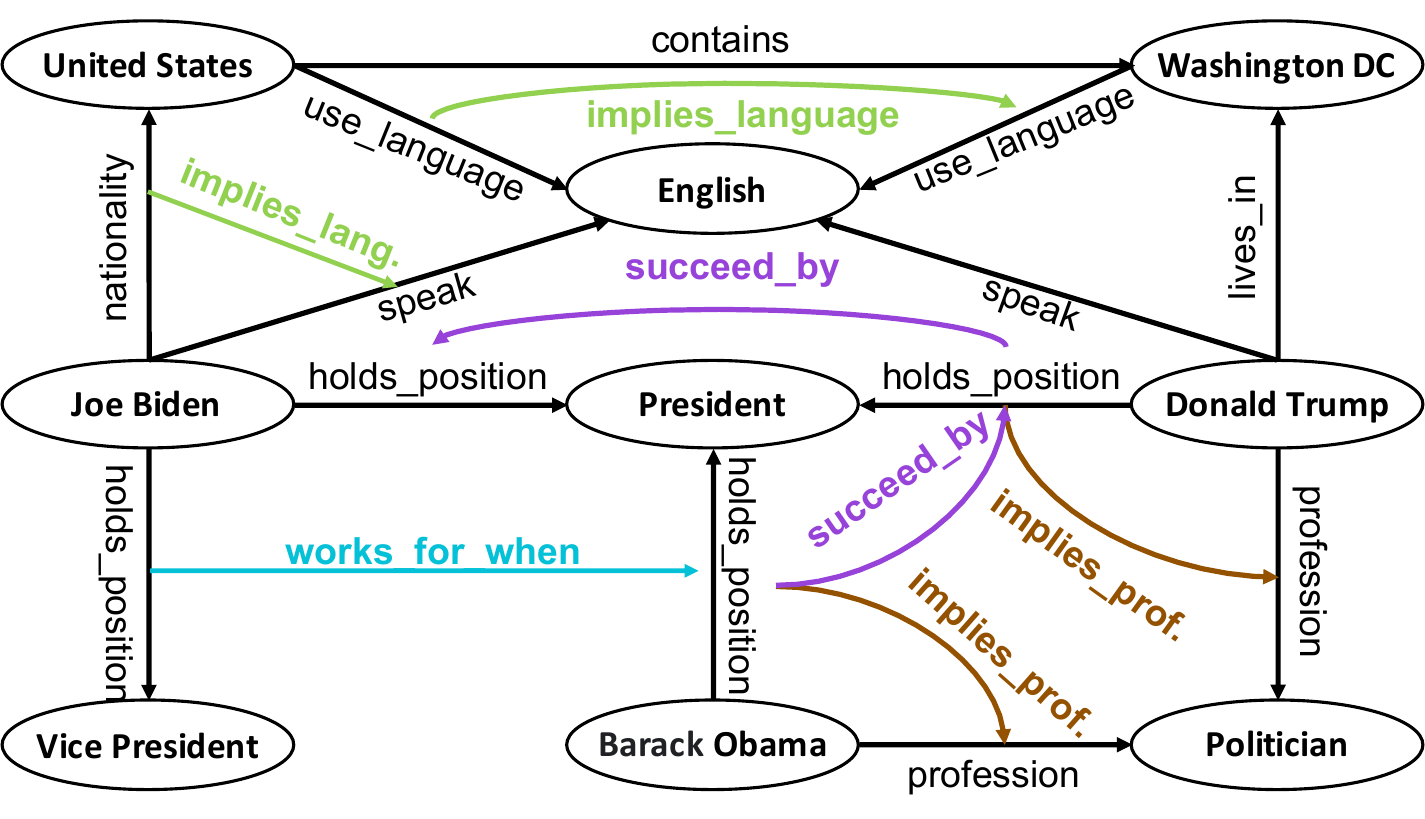}
\vspace{-0.2cm}
\caption{An example of a nested factual KG consisting of 1) a set of atomic facts describing the relationship between entities and 2) a set of nested facts describing the relationship between atomic facts. Nested factual relations are colored and they either describe situations in/over time (e.g., \emph{succeed\_by} and \emph{works\_for\_when}) or logical patterns (e.g., \emph{implies\_profession} and \emph{implies\_language}). 
}
\label{fig:bikg}
\vspace{-0.6cm}
\end{figure}

However, these beyond-triple representations typically focus only on relationships between entities that jointly define an \emph{atomic fact}, overlooking the significance of relationships that describe multiple facts together.
Indeed, within a KG, each atomic fact may have a relationship with another atomic fact. Consider the following two atomic facts:
$T_1$=(\emph{JoeBiden}, \emph{holds\_position}, \emph{VicePresident}) and $T_2$=(\emph{BarackObama}, \emph{holds\_position}, \emph{President}).
We can depict the scenario where \emph{JoeBiden} held the position of \emph{VicePresident} under the \emph{President} \emph{BarackObama} using a triple $(T_1, \emph{works\_for\_when}, T_2)$.
Such a fact about facts is referred to as a \emph{nested fact} \footnote{This is also called a \emph{quoted triple} in RDF star \cite{DBLP:journals/ercim/Champin22}.}
and the relation connecting these two facts is termed a \emph{nested relation}.
Fig. \ref{fig:bikg} provides an illustration of a KG containing both atomic and nested facts.

These nested relations play a crucial role in expressing complex semantics and queries in two ways:
1) \textbf{Expressing situations involving facts in or over time}. This facilitates answering complex queries that involve multiple facts.
For example, KG embeddings face challenges when addressing queries like "\emph{Who was the president of the USA after \emph{DonaldTrump}?}" because the query about the primary fact (?, \emph{holds\_position}, \emph{President}) depends on another fact (\emph{DonaldTrump}, \emph{holds\_position}, \emph{President}). As depicted in Fig. \ref{fig:bikg}, \emph{succeed\_by} conveys such temporal situation between these two facts, allowing the direct response to the query through conditional link prediction.
2) \textbf{Expressing logical patterns (implications) using a non-first-order logical form $\psi \stackrel{\widehat{r}}{\rightarrow}\phi$}.
As illustrated in Fig. \ref{fig:bikg}, (\emph{Location A}, \emph{uses\_language}, \emph{Language B}) $\stackrel{\emph{implies\_language}}{\rightarrow} $ (\emph{Location in A}, \emph{uses\_language}, \emph{Language B}) represents a logical pattern, as it holds true for all pairs of (\emph{Location A}, \emph{Location in A}).
Modeling such logical patterns is crucial as it facilitates generalization. Once these patterns are learned, new facts adhering to these patterns can be inferred.
A recent study \cite{DBLP:conf/aaai/Chanyoung} explored link prediction over nested facts.\footnote{In their work, the KG is referred to as a bi-level KG, and the term "high-level facts" is synonymous with nested facts.}
However, their method embeds facts using a multilayer perceptron (MLP), which fails to capture essential logical patterns and thus has limited generalization capabilities.

In this paper, we introduce NestE, an innovative approach designed to embed the semantics of both atomic facts and nested facts that enable representing temporal situations and logical patterns over facts. 
NestE represents each atomic fact as a $1\times3$ hypercomplex matrix, with each element signifying a component of the atomic fact. Furthermore, each nested relation is modeled through a $3\times3$ hypercomplex matrix that rotates the $1\times3$ atomic fact matrix via a matrix-multiplicative Hamilton product. 
Our matrix-like modeling for facts and nested relations demonstrates the capacity to encode diverse logical patterns over nested facts. The modeling of these logical patterns further enables efficient modeling of logical rules that extend beyond the first-order-logic-like expressions (e.g., Horn rules).
Moreover, we propose a more general hypercomplex embedding framework that extends the quaternion embedding \cite{DBLP:conf/nips/0007TYL19} to include hyperbolic quaternions and split-quaternions. This generalization of hypercomplex space allows for expressing rotations over hyperboloid, providing more powerful and distinct inductive biases for embedding complex structural patterns (e.g., hierarchies). 
Our experimental findings on triple prediction and conditional link prediction showcase the remarkable performance gain of NestE.

\section{Related Work}

\paragraph{Beyond-Triple KGs} 
To enrich the semantics of triple-base KGs, several lines of work have explored more powerful representations \cite{DBLP:conf/cikm/XiongNDC23}. 
Temporal KGs \cite{DBLP:conf/emnlp/DasguptaRT18,DBLP:conf/www/LeblayC18,DBLP:journals/corr/abs-2308-02457} introduce an additional timestamp to each triple to specify the temporal validity of the fact. 
Hyper-relational facts \cite{DBLP:conf/acl/GuanJGWC20, DBLP:conf/www/RossoYC20, DBLP:conf/emnlp/GalkinTMUL20,xiong2023shrinking} attach a set of key-value qualifiers to the primal triple, where each qualifier specify certain semantics of the primal fact. 
$N$-ary facts \cite{DBLP:conf/www/0016Y020,DBLP:conf/www/LiuYL21,DBLP:conf/ijcai/FatemiTV020} represent a fact as an abstracted relationship between $n$ entities.
Bilinear models are generalized to n-ary facts by replacing the bilinear product with multi-linear products \cite{DBLP:conf/www/0016Y020,DBLP:conf/www/LiuYL21}. 
These representations capture relationships between entities or between entities and facts, but they do not capture the relationships between multiple facts.

\paragraph{Describing relationships between facts} 
Rule-based approaches \cite{DBLP:conf/aaai/NiuZ0CLLZ20,DBLP:conf/ijcai/MeilickeCRS19,DBLP:conf/emnlp/DemeesterRR16,DBLP:conf/emnlp/GuoWWWG16,DBLP:conf/nips/YangYC17,DBLP:conf/nips/SadeghianADW19} consider relationships between facts, but they are confined to first-order-logic-like expressions (i.e., Horn rules), i.e., $\forall e_1,e_2,e_3: (e_1,r_1,e_2)\wedge (e_2,r_2,e_3) \Rightarrow (e_1,r_3,e_3)$, where there must exist a path connecting $e_1$, $e_2$, and $e_3$ in the KG. 
Notably, \cite{DBLP:conf/aaai/Chanyoung} marked an advancement by examining KG embeddings with relationships between facts as nested facts, denoted as $(x, r_1, y) \xRightarrow{\hat{r}} (p, r_2, q)$. The proposed embeddings (i.e., BiVE-Q and BiVE-B\footnote{Note that BiVE-B, despite being described as based on the biquaternian--BiQUE, employs quaternion space with an additional translation component based on our analysis of the code.}) concatenate the embeddings of the head, relation, and tail, subsequently embedding them via an MLP. However, such modeling does do not explicitly capture crucial logical patterns over nested facts, which bear significant importance in KG embeddings.

\paragraph{Algebraic and geometric embeddings} 
Algebraic embeddings like QuatE \cite{DBLP:conf/nips/0007TYL19} and BiQUE \cite{DBLP:conf/emnlp/GuoK21} represent relations as algebraic operations and score triples using inner products. They can be viewed as a unification of many earlier functional \cite{DBLP:conf/nips/BordesUGWY13} and multiplication-based \cite{DBLP:conf/icml/TrouillonWRGB16} models. 
Geometric embeddings like hyperbolic embeddings \cite{DBLP:conf/acl/ChamiWJSRR20,DBLP:conf/nips/BalazevicAH19} further extend the functional models to non-Euclidean hyperbolic space, enabling the representation of hierarchical relations.

\section{Preliminaries}

A KG is denoted as a graph $G=(\sV,\sR,\sT)$, where $\sV$ represents the set of entities, $\sR$ stands for the set of relation names, and $\sT=\{(h,r,t) : h,t\in\sV, r\in\sR\}$ represents the set of triples. We refer to $G$ as an atomic factual KG, and each $(h,r,t)\in\sT$ is referred to as an atomic triple. 
The nested triple and nested factual KG are defined as follows.

\begin{mydef}[Nested Triple]
Given an atomic factual KG $G=(\sV,\sR,\sT)$, a set of nested triples is defined by $\sTT=\{\langle T_i, \rhat, T_j\rangle:T_i,T_j\in\sT, \rhat\in\Rhat \}$, where $\sT$ is the set of atomic triples and $\Rhat$ is the set of nested relation names.
\end{mydef}


\begin{mydef}[Nested Factual Knowledge Graph]
Given a KG $G=(\sV,\sR,\sT)$, a set of nested relation names $\Rhat$, and a set of nested triples $\sTT$ defined on $G$ and $\Rhat$,
a nested factual KG is defined as $\Ghat=(\sV, \sR, \sT, \Rhat, \sTT)$.
\end{mydef}

We can now define triple prediction and conditional link prediction \cite{DBLP:conf/aaai/Chanyoung} as follows.

\begin{mydef}[Triple Prediction]
Given a nested factual KG $\Ghat=(\sV, \sR, \sT, \Rhat, \sTT)$, the triple prediction problem involves answering a query $\langle T_i, \rhat, ?t\rangle$ or $\langle h? , \rhat, T_j\rangle$ with $T_i, T_j \in \sT$ and $\rhat \in \Rhat$, where the variable $?h$ or $?t$ needs to be bounded to an atomic triple within $\Ghat$.
\end{mydef}

\begin{mydef}[Conditional Link Prediction]
Given a nested factual KG $\Ghat=(\sV, \sR, \sT, \Rhat, \sTT)$, let $T_i = (h_i,r_i,t_i)$ and $T_j = (h_j,r_j,t_j)$. The conditional link prediction problem involves queries $\langle T_i, \rhat, (h_j,r_j,?) \rangle$, $\langle T_i, \rhat, (?,r_j,t_j) \rangle$, $\langle (h_i,r_i,?), \rhat, T_j \rangle$, or $\langle (?,r_i,t_i), \rhat, T_j \rangle$, where the variables need to be bound to entities within $\Ghat$.
\end{mydef}

\section{NestE: Embedding Atomic and Nested Facts }

\subsection{Unified Hypercomplex Embeddings}

We first extend QuatE \cite{DBLP:conf/nips/0007TYL19}, a KG embedding in 4D hypercomplex quaternion space, into a more general 4D hypercomplex number system including three variations: (spherical) quaternions, hyperbolic quaternions, and split quaternions. Each of these 4D hypercomplex numbers is composed of one real component and three imaginary components denoted by $s+x\textit{i}+y\textit{j}+z\textit{k}$ with $s,x,y,z \in \mathbb{R}$ and $i,j,k$ being the three imaginary parts. 
The distinctive feature among these hyper-complex number systems lies in their multiplication rules of the imaginary components.

\noindent\textbf{(Spherical) quaternions $\mathcal{Q}$} follow the multiplication rules:
\begin{equation}
\begin{array}{l}
\textit{i}^2 = \textit{j}^2 = \textit{k}^2 = 1, \\
\textit{ij} = \textit{k}=-ji, \textit{jk} = \textit{i}=-kj, \textit{ki} = \textit{j}=-ik. \
\end{array}
\end{equation}

\noindent\textbf{Hyperbolic quaternions $\mathcal{H}$} follow the multiplication rules:
\begin{equation}
\begin{array}{l}
\textit{i}^2 = -1, \textit{j}^2 = \textit{k}^2 = 1, \\
\textit{ij} = \textit{k}, \textit{jk} = -\textit{i}, \textit{ki} = \textit{j}, \
\textit{ji} = -\textit{k}, \textit{kj} = \textit{i}, \textit{ik} = -\textit{j}.
\end{array}
\end{equation}

\noindent\textbf{
Split quaternions $\mathcal{S}$} follow the multiplication rules:
\begin{equation}
\begin{array}{l}
\textit{i}^2 = -1, \textit{j}^2 = \textit{k}^2 = 1, \\
\textit{ij} = \textit{k}, \textit{jk} = -\textit{i}, \textit{ki} = \textit{j}, \
\textit{ji} = -\textit{k}, \textit{kj} = \textit{i}, \textit{ik} = -\textit{j}.
\end{array}
\end{equation}

\paragraph{Geometric intuitions}
The distinctions in the multiplication rules of various hypercomplex numbers give rise to different geometric spaces that provide suitable inductive biases for representing different types of relations. Specifically, spherical quaternions, hyperbolic quaternions, and split quaternions with the same norm $c$ correspond to 4D hypersphere, Lorentz model of hyperbolic space (i.e., the upper part of the double-sheet hyperboloid), and pseudo-hyperboloid (i.e., one-sheet hyperboloid, with curvature $\sqrt{c}$, respectively. These are denoted as follows:
\begin{equation}\small
\begin{array}{l}
|\mathcal{Q}| = s^2 + x^2 + y^2 + z^2 = c > 0 \ (\text{hypersphere}) \\
|\mathcal{H}| = s^2 - x^2 - y^2 - z^2 = c > 0 \ (\text{Lorentz hyperbolic space}) \\
|\mathcal{S}| = s^2 + x^2 - y^2 - z^2 = c > 0 \ (\text{pseudo-hyperboloid}). \\
\end{array}
\end{equation}
These spaces have well-known characteristics: spherical spaces are adept at modeling cyclic relations \cite{DBLP:conf/www/WangWSWNAXYC21}, hyperbolic spaces provide geometric inductive biases for hierarchical relations \cite{DBLP:conf/acl/ChamiWJSRR20}, and the pseudo-hyperboloid \cite{DBLP:conf/kdd/XiongZNXP0S22} offers a balance between spherical and hyperbolic spaces, making it suitable for embedding both cyclic and hierarchical relations.
Moreover, by representing relations as geometric rotations over these spaces (i.e., Hamilton product), fundamental logical patterns such as symmetry, inversion, and compositions can be effectively inferred \cite{DBLP:conf/nips/0007TYL19,DBLP:conf/acl/ChamiWJSRR20,DBLP:conf/kdd/XiongZNXP0S22}. 
Our proposed embeddings can be viewed as a unification of previous approaches that leverages these geometric inductive biases in these geometric spaces within a single geometric algebraic framework.

For convenience, we parameterize each entity and relation as a Cartesian product of $d$ 4D hypercomplex numbers $\mathbf{s} + \mathbf{x}\textit{i} + \mathbf{y}\textit{j} + \mathbf{z}\textit{k}$, where $\mathbf{s}, \mathbf{x}, \mathbf{y}, \mathbf{z} \in \mathbb{R}^d$.
This enables us to define all algebraic operations involving these hypercomplex vectors in an element-wise manner.


\subsection{Atomic Fact Embeddings}

Each atomic relation is represented by a rotation hypercomplex vector $\mathbf{r}_\theta$ and a translation hypercomplex vector $\mathbf{r}_b$. For a given triple $(h, r, t)$, we apply the following operation:
\begin{equation}
\mathbf{h}^\prime = ( \mathbf{h} \oplus \mathbf{r}_b ) \otimes \mathbf{r}_\theta,
\end{equation}
where $\oplus$ and $\otimes$ stand for addition and Hamilton product between hypercomplex numbers, respectively. The addition involves an element-wise sum of each hypercomplex component. The Hamilton product rotates the head entity. To ensure proper rotation on the unit sphere, we normalize the rotation hypercomplex number $\mathbf{r}_\theta = \mathbf{s}_r^\theta+\mathbf{x}_r^\theta\textit{i}+\mathbf{y}_r^\theta \textit{j}+\mathbf{z}_r^\theta \textit{k}$ by $\mathbf{r}_\theta=\frac{\mathbf{s}_r^\theta+\mathbf{x}_r^\theta \boldsymbol{i}+\mathbf{y}_r^\theta \boldsymbol{j}+\mathbf{z}_r^\theta \boldsymbol{k}}{\sqrt{ {\mathbf{s}_r^\theta}^2+{\mathbf{x}_r^\theta}^2+{\mathbf{y}_r^\theta}^2+{\mathbf{z}_r^\theta}^2}}$. 
Hamilton product is defined by combining the components of the hypercomplex numbers. 
\begin{equation}
\begin{split}
& \mathbf{h}^\prime = \mathbf{h} \otimes \mathbf{r}_\theta \\
& =\left(\mathbf{s}_h \circ \mathbf{s}_r^\theta \circ 1 + \mathbf{x}_h \circ \mathbf{x}_r^\theta \circ i^2  + \mathbf{y}_h \circ \mathbf{y}_r^\theta  \circ j^2 + \mathbf{z}_h \circ \mathbf{z}_r^\theta \circ k^2 \right) \\
& +\left(\mathbf{s}_h \circ \mathbf{x}_r^\theta \circ i + \mathbf{x}_h \circ \mathbf{s}_r^\theta  \circ i  + \mathbf{y}_h \circ \mathbf{z}_r^\theta \circ jk + \mathbf{z}_h \circ \mathbf{y}_r^\theta \circ kj \right) \\
& +\left(\mathbf{s}_h \circ \mathbf{y}_r^\theta \circ j + \mathbf{x}_h \circ \mathbf{z}_r^\theta \circ ik + \mathbf{y}_h \circ \mathbf{s}_r^\theta \circ j  + \mathbf{z}_h \circ \mathbf{x}_r^\theta \circ ik \right)  \\
& +\left(\mathbf{s}_h \circ \mathbf{z}_r^\theta \circ k + \mathbf{x}_h \circ \mathbf{y}_r^\theta \circ ij + \mathbf{y}_h \circ \mathbf{x}_r^\theta \circ ij + \mathbf{z}_h \circ \mathbf{s}_r^\theta \circ k \right) \\ 
& = \mathbf{s}_{h^\prime}+\mathbf{x}_{h^\prime}\textit{i}+\mathbf{y}_{h^\prime} \textit{j}+\mathbf{z}_{h^\prime} \textit{k},
\end{split}
\end{equation}
where the multiplication of imaginary components follows the rules (Eq.1-3) of the chosen hypercomplex systems.



The scoring function $\phi(h, r, t)$ is defined as:
\begin{equation}\small
\phi(h, r, t)= \langle \mathbf{h}^\prime, \mathbf{t} \rangle =\left\langle \mathbf{s}_{h^\prime}, \mathbf{s}_t\right\rangle+\left\langle \mathbf{x}_{h^\prime}, \mathbf{x}_t\right\rangle+\left\langle \mathbf{y}_{h^\prime}, \mathbf{y}_t\right\rangle+\left\langle \mathbf{z}_{h^\prime}, \mathbf{z}_t\right\rangle,
\end{equation}
where $\langle \cdot, \cdot \rangle$ represents the inner product. 

\subsection{Nested Fact Embeddings}

To represent an atomic fact $(h, r, t)$ without losing information, we embed each atomic triple as a $1\times3$ matrix, where each column corresponds to the embedding of the respective element. Consequently, we have $\mathbf{T}_i = \left[\mathbf{h}_i,\mathbf{r}_i,\mathbf{t}_i\right]$.

To embed various shapes of nested relations between $T_i$ and $T_j$ with relation $\widehat{r}$, 
we model each nested relation using a $1\times3$ translation matrix and a $3\times3$ rotation matrix, where each element is a 4D hypercomplex number. Specifically, we first translate the head triple $\mathbf{T}_i$ with $\mathbf{\widehat{r}}_b$, followed by applying a matrix-like rotation of $\mathbf{\widehat{r}}_\theta$, as defined by
\begin{equation}
\mathbf{T}i^\prime = ( \mathbf{T}i \oplus_{1 \times 3} \mathbf{\widehat{r}}_b ) \otimes_{3 \times 3} \mathbf{\widehat{r}}_\theta,
\end{equation}
where the matrix addition $\oplus_{1 \times 3}$ is performed through an element-wise summation of the hypercomplex components within the matrices. The matrix-like Hamilton product $\otimes_{3 \times 3}$ is defined as a product akin to matrix multiplication:
\begin{equation}\small
\begin{split}
\mathbf{T}_i^\prime =  \mathbf{T}_i \otimes_{3 \times 3} \mathbf{ \widehat{r} }_\theta = 
\left[
\begin{array}{c}
   \mathbf{h}_i\\
   \mathbf{r}_i\\
   \mathbf{t}_i\\
\end{array}
\right]^\top \times
\left[ 
\begin{array}{ccc}
   \mathbf{\widehat{r}}_{11}^\theta & \mathbf{\widehat{r}}_{12}^\theta & \mathbf{\widehat{r}}_{13}^\theta\\
   \mathbf{\widehat{r}}_{21}^\theta & \mathbf{\widehat{r}}_{22}^\theta & \mathbf{\widehat{r}}_{23}^\theta\\
   \mathbf{\widehat{r}}_{31}^\theta & \mathbf{\widehat{r}}_{32}^\theta & \mathbf{\widehat{r}}_{33}^\theta\\
\end{array}
\right] = \\
\left[
\begin{array}{c}
   \mathbf{h}_i \otimes \mathbf{\widehat{r}}_{11}^\theta + \mathbf{r}_i \otimes \mathbf{\widehat{r}}_{21}^\theta + \mathbf{t}_i \otimes \mathbf{\widehat{r}}_{31}^\theta   \\
   \mathbf{h}_i \otimes \mathbf{\widehat{r}}_{12}^\theta + \mathbf{r}_i \otimes \mathbf{\widehat{r}}_{22}^\theta + \mathbf{t}_i \otimes \mathbf{\widehat{r}}_{32}^\theta   \\
   \mathbf{h}_i \otimes \mathbf{\widehat{r}}_{13}^\theta + \mathbf{r}_i \otimes \mathbf{\widehat{r}}_{23}^\theta + \mathbf{t}_i \otimes \mathbf{\widehat{r}}_{33}^\theta   \\
\end{array}
\right]^\top = 
\left[
\begin{array}{c}
   \mathbf{h}_i^\prime\\
   \mathbf{r}_i^\prime\\
   \mathbf{t}_i^\prime\\
\end{array}
\right]^\top,
\end{split}
\end{equation}
where $\otimes$ is the Hamilton product. 

\paragraph{Remarks} 
This matrix-like modeling of nested facts provides flexibility to capture diverse shapes of logical patterns inherent in nested relations. In essence, different shapes of situations or patterns can be effectively modeled by manipulating the $3\times3$ rotation matrix. For instance, relational implications can be represented using a diagonal matrix, while inversion can be captured using an anti-diagonal matrix. See \textbf{theoretical justification} for further analysis.

To assess the plausibility of the nested fact $(T_i, \widehat{r}, T_j)$, we calculate the inner product between the transformed head $\mathbf{T}_i^\prime$ fact and the tail fact $\mathbf{T}_j$ as:
\begin{equation}
\rho(T_i, \widehat{r}, T_j)= \langle\mathbf{T}_i^\prime, \mathbf{T}_j \rangle,
\end{equation}
where $\langle \cdot, \cdot \rangle$ denotes the matrix inner product.

\paragraph{Learning objective}

We sum up the loss of atomic fact embedding $\mathcal{L}_{\text{atomic}}$, the loss of nested fact embedding $\mathcal{L}_{\text{meta}}$, and additionally the loss term $\mathcal{L}_{\text{aug}}$ for augmented triples generated by random walking as used in \cite{DBLP:conf/aaai/Chanyoung}. The overall loss is defined as
\begin{equation}
    \mathcal{L} = \mathcal{L}_{\text{atomic}} + \lambda_1 \mathcal{L}_{\text{nested}} + \lambda_2 \mathcal{L}_{\text{aug}}
\end{equation}
where $\lambda_1$ and $\lambda_2$ are the weight hyperparameters indicating the importance of each loss. 
Negative sampling is applied by randomly replacing one of the head or tail entity/triple. 
These losses are defined as follows:
\begin{equation}\footnotesize 
    \begin{array}{@{} *{6}{>{\displaystyle}c} @{}}
    \mathcal{L}_{\text{atomic}} = \sum_{(h,r,t) \in \mathcal{T}} g\left(-\phi\left(h,r,t\right) \right) + \sum_{g\left(\left(h^\prime,r^\prime,t^\prime\right)\right) \notin \mathcal{T}} g\left(\phi\left(h^\prime,r^\prime,t^\prime\right)\right)\\
     \mathcal{L}_{\text{aug}} = \sum_{(h,r,t) \in \mathcal{T}^\prime}g\left(-\phi\left(h,r,t\right)\right) + \sum_{(h^\prime,r^\prime,t^\prime) \notin \mathcal{T}^\prime}g\left(\phi(h^\prime,r^\prime,t^\prime)\right)\\
    \mathcal{L}_{\text{nested}} = \sum_{(T_i,\widehat{r},T_j) \notin \sTT } g\left(-\rho(T_i,\widehat{r},T_j)\right) + \sum_{(T_i^\prime,\widehat{r},T_j^\prime) \notin \sTT} g\left(\rho(T_i^\prime,\widehat{r},T_j^\prime)\right),\\
    \end{array}
\end{equation}
where $g=\log(1+\exp(x))$ and $\mathcal{T}^\prime$ is the set of augmented triples. 

\subsection{Theoretical Justification}
\label{sec:theory}

Modeling logical patterns is of great importance for KG embeddings because it enables generalization, i.e., once the patterns are learned, new facts that respect the patterns can be inferred. A logical pattern is a logical form $\psi \rightarrow \phi$ with $\psi$ and $\phi$ being the body and head, implying that if the body is satisfied then the head must also be satisfied. 

\paragraph{First-order-logic-like logical patterns} 
Existing KG embeddings studied logical patterns expressed in the first-order-logic-like form. 
Prominent examples include symmetry $\forall h,t\colon (h,r,t) \rightarrow (t,r,h)$, anti-symmetry $\forall h,t\colon (h,r,t) \rightarrow \neg (h,r,t)$, inversion $\forall h,t\colon (h,r_1,t) \rightarrow (t,r_2,h)$ and composition $\forall e_1,e_2,e_3\colon (e_1,r_1,e_2) \wedge (e_2,r_2,e_3)  \rightarrow (e_1,r_3,e_3)$.

\begin{proposition}
NestE can infer symmetry, anti-symmetry, inversion, and composition, regardless of the specific choices of hypercomplex number systems.
\end{proposition}
This proposition holds because NestE subsumes ComplEx \cite{DBLP:conf/icml/TrouillonWRGB16} (i.e., 4D complex numbers generalize 2D complex numbers).

\begin{figure}
    \centering
    \includegraphics[width=\linewidth]{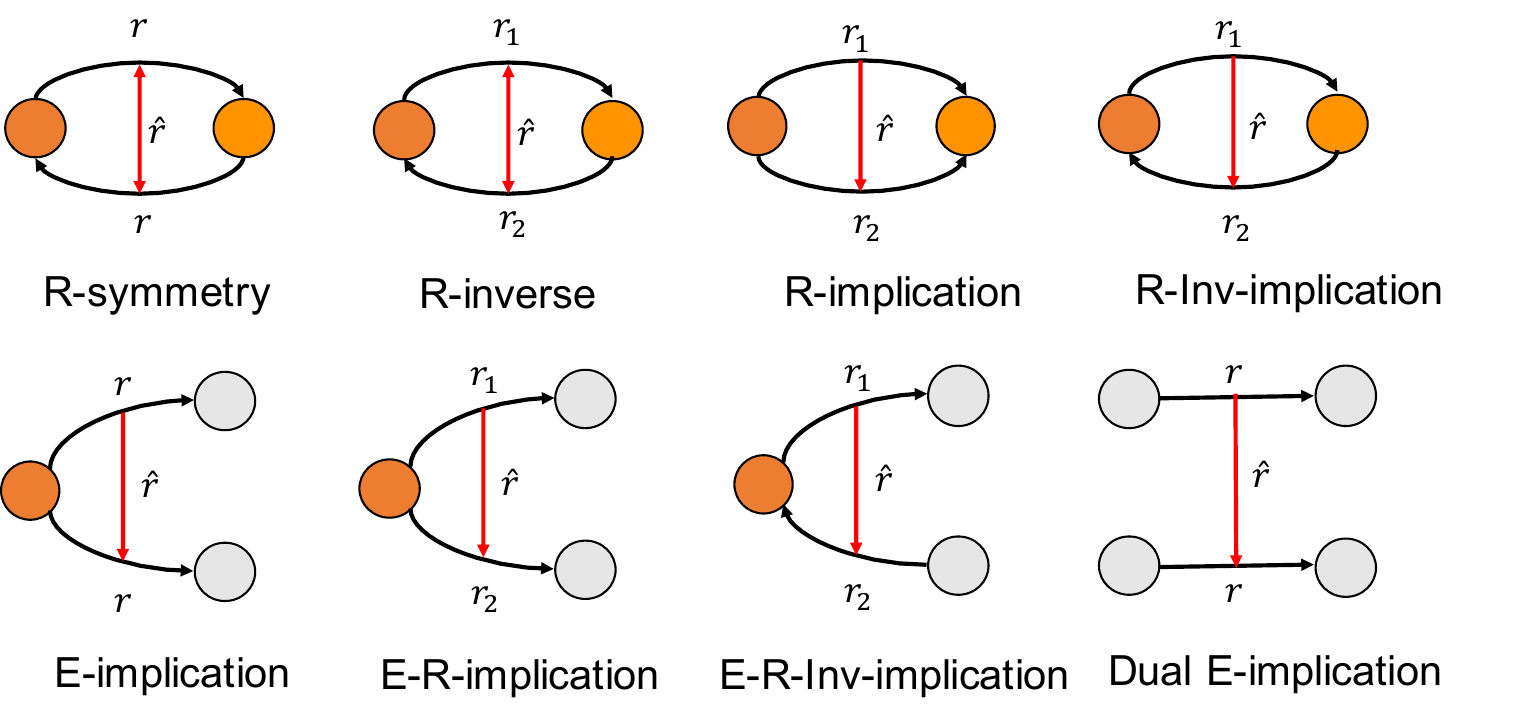}
    \vspace{-0.6cm}
    \caption{A structural illustration of different shapes of logical patterns, where the colored circles are free variables. 
    }
    \vspace{-0.1cm}
    \label{fig:pattern}
\end{figure}



\paragraph{Logical patterns over nested facts}
We extend the vanilla logical patterns in KGs to include nested facts. 
This can be expressed in a non-first-order-logic-like form $\psi \stackrel{\widehat{r}}{\rightarrow}\phi$.

\begin{itemize}
    \item \textbf{Relational symmetry (R-symmetry):} an atomic relation $r$ is symmetric w.r.t a nested relation $\widehat{r}$ if $\forall x,y \in \mathcal{E},  \langle x, r, y\rangle  \stackrel{\widehat{r}}{\leftrightarrow} \langle y, r, x\rangle$. 

    \item \textbf{Relational inverse (R-inverse):} two atomic relations $r_1$ and $r_2$ are inverse w.r.t a nested relation $\widehat{r}$ if $\forall x,y \in \mathcal{E}, $ $(\langle x, r_1, y\rangle \stackrel{\widehat{r}}{\leftrightarrow} \langle y, r_2, x\rangle)$.

    \item \textbf{Relational implication (R-implication):} an atomic relation $r_1$ implies a atomic relation $r_2$ w.r.t a nested relation $\widehat{r}$ if $\forall x,y \in \mathcal{E}, (\langle x, r_1, y\rangle \stackrel{\widehat{r}}{\rightarrow} \langle x, r_2, y\rangle)$.

    \item \textbf{Relational inverse implication (R-Inv-implication):} an atomic relation $r_1$ inversely implies an atomic relation $r_2$ w.r.t a nested relation $\widehat{r}$ if $\forall x,y \in \mathcal{E}, (\langle x, r_1, y\rangle \stackrel{\widehat{r}}{\rightarrow} \langle y, r_2, x\rangle)$.

    \item \textbf{Entity implication (E-implication):} an entity $x_1$ (resp. $y_1$) implies entity $x_2$ (resp. $y_2$) w.r.t an atomic relation $r$ and a nested relation $\widehat{r}$ if  $\forall y \in \mathcal{E}, (\langle x_1, r, y\rangle \stackrel{\widehat{r}}{\rightarrow} \langle x_2, r, y\rangle)$ (resp. $\forall x \in \mathcal{E}, (\langle x, r, y_1\rangle \stackrel{\widehat{r}}{\rightarrow} \langle x, r, y_2\rangle)$ ).

    \item \textbf{Entity relational implication (E-R-implication):} an entity $x_1$ and relation $r_1$ (resp. $y_1$ and relation $r_1$) implies entity $x_2$ and relation $r_2$ (resp. $y_2$ and relation $r_2$) 
    w.r.t a nested relation $\widehat{r}$ if  $\forall y \in \mathcal{E}, (\langle x_1, r_1, y\rangle \stackrel{\widehat{r}}{\rightarrow} \langle x_2, r_2, y\rangle)$ (resp. $\forall x \in \mathcal{E}, (\langle x, r_1, y_1\rangle \stackrel{\widehat{r}}{\rightarrow} \langle x, r_2, y_2\rangle)$).

    \item \textbf{Entity relational inverse implication (E-R-Inv-implication):}  an entity $x_1$ and relation $r_1$ (resp. $y_1$ and relation $r_1$) inversely implies entity $x_2$ and relation $r_2$ (resp. $y_2$ and relation $r_2$) w.r.t a nested relation $\widehat{r}$ if  $\forall y \in \mathcal{E}, (\langle x_1, r_1, y\rangle \stackrel{\widehat{r}}{\rightarrow} \langle y, r_2, x_2\rangle)$ (resp. $\forall x \in \mathcal{E}, (\langle x, r_1, y_1\rangle \stackrel{\widehat{r}}{\rightarrow} \langle y_2, r_2, x\rangle)$).
     \item \textbf{Dual Entity implication (Dual E-implication):} an entity pair ($x_1$, $x_2$) implies another entity pair ($y_1$, $y_2$)  iff both ($x_1$, $y_1$) and ($x_2$, $y_2$) satisfy E-implication. 

\end{itemize}

Fig. \ref{fig:pattern} illustrates the structure of the introduced patterns and Table \ref{tab:pattern} presents exemplary patterns of nested facts.

\begin{table}[]
    \centering
    \setlength{\tabcolsep}{0.1em}
     \caption{Exemplary logical patterns of nested facts.  }
    \resizebox{\linewidth}{!}{
    \begin{tabular}{c|ccccc}
        Pattern & $\widehat{r}$ & triple template & \\
        \midrule
        \multirow{2}{*}{R-Symmetry} & \multirow{2}{*}{EquivalentTo} & (Person A, \underline{IsMarriedTo}, Person B)  \\
         & & (Person B, \underline{IsMarriedTo}, Person A) \\
        \multirow{2}{*}{R-Inverse} & \multirow{2}{*}{EquivalentTo} & (Location A, \underline{UsesLanguage}, Language B)  \\
         & & (Language B, \underline{IsSpokenIn}, Location A) \\
         \multirow{2}{*}{R-Implication} & \multirow{2}{*}{ImpliesLocation} & (Country A, \underline{CapitalIsLocatedIn}, City B)  \\
         & & (Country A, \underline{Contains}, City B) \\
           \multirow{2}{*}{R-Inv-Implication} & \multirow{2}{*}{ImpliesLocation} & (Organization A, \underline{Headquarter}, Location B)  \\
         & & (Location B, \underline{Contains}, Organization A) \\
        \multirow{2}{*}{E-Implication} & \multirow{2}{*}{ImpliesTimeZone} & (\underline{Location A}, TimeZone, Time Zone B)  \\
         & & (\underline{Location in A}, TimeZone, Time Zone B) \\
         \multirow{2}{*}{E-R-Implication} & \multirow{2}{*}{ImpliesProfession} & (Person A, \underline{HoldsPosition, Government Position B})  \\
         & & (Person A, \underline{IsA, Politician}) \\
          \multirow{2}{*}{E-R-Inv-Implication} & \multirow{2}{*}{ImpliesProfession} & (\underline{Work A, CinematographyBy}, Person B)  \\
         & & (Person B, \underline{IsA, Cinematographer}) \\
           \multirow{2}{*}{Dual E-Implication} & \multirow{2}{*}{ImpliesLocation} & (\underline{Location A}, Contains, \underline{Location B})  \\
         & & (\underline{Location containing A}, Contains, \underline{Location in B}) \\
        \bottomrule
    \end{tabular}
    }
    \label{tab:pattern} 
    \vspace{-0.3cm}
\end{table}

\begin{proposition}
    NestE can infer R-symmetry, R-inverse, R-implication, R-Inv-implication, E-implication, E-R-implication, E-R-Inv-implication, and Dual E-implication.
\end{proposition}

\begin{hproof}
To infer different logical patterns via different free variables, we can set some elements of the relation matrix to be zero-valued or one-valued complex numbers. For example, the implication and inverse implication relations can be inferred by setting the matrix to be diagonal or anti-diagonal. See Appendix for details.
\end{hproof}


\section{Experimental Results}

\subsection{Experiment Setup}

\paragraph{Datasets} 
We utilize three benchmark KGs: FBH, FBHE, and DBHE, that contain nested facts and are constructed by \cite{DBLP:conf/aaai/Chanyoung}. 
FBH and FBHE are based on FB15K237 from Freebase \cite{DBLP:conf/sigmod/BollackerEPST08} while DBHE is based on DB15K from DBpedia \cite{DBLP:conf/semweb/AuerBKLCI07}. 
FBH contains only nested facts that can be inferred from the triple facts, e.g., \emph{prerequisite\_for} and \emph{implies\_position}, while FBHE and DBHE further contain externally-sourced knowledge crawled from Wikipedia articles, e.g., \emph{next\_almaMater} and \emph{transfers\_to}. 
The authors of \cite{DBLP:conf/aaai/Chanyoung} spent six weeks manually defining these nested facts and adding them to the KGs. 
The dataset details are presented in Table~\ref{tb:data}. We split $\sT$ and $\sTT$ into training, validation, and test sets in an 8:1:1 ratio. 

\begin{table}[t]
\small
\centering
\setlength{\tabcolsep}{0.65em}
\caption{Statistics of $\Ghat=(V, R, \sT, \Rhat, \sTT)$. $|\sT|^\prime$ denotes the number of atomic triples involved in the nested triples.}
\vspace{-0.3cm}
\begin{tabular}{ccccccc}
\toprule
 & $|V|$ & $|R|$ & $|\sT|$ & $|\Rhat|$ & $|\sTT|$ & $|\sT|^\prime$ \\
\midrule
\fbh & 14,541 & 237 & 310,117 & 6 & 27,062 & 33,157 \\
\fbhe & 14,541 & 237 & 310,117 & 10 & 34,941 & 33,719 \\
\dbhe & 12,440 & 87 & 68,296 & 8 & 6,717 & 8,206 \\
\bottomrule
\end{tabular}
\label{tb:data}
 \vspace{-0.4cm}
\end{table}

\paragraph{Baselines} 
We consider BiVE-Q and BiVE-B \cite{DBLP:conf/aaai/Chanyoung} as our major baselines as they are specifically designed for KGs with nested facts and have demonstrated significant improvements over triple-based methods. 
We also compare some rule-based approaches as they indirectly consider relations between facts in first-order-logic-like expression, including Neural-LP \cite{DBLP:conf/nips/YangYC17}, DRUM \cite{DBLP:conf/nips/SadeghianADW19}, and AnyBURL \cite{DBLP:conf/ijcai/MeilickeCRS19}.
We further include QuatE \cite{DBLP:conf/nips/0007TYL19} and BiQUE \cite{DBLP:conf/emnlp/GuoK21} as they are the SoTA triple-based methods and they are also based on 4D hypercomplex numbers. 
However, these triple-based methods do not directly apply to the nested facts. 
Following \cite{DBLP:conf/aaai/Chanyoung}, we create a new triple-based KG $G_T$ where the atomic facts are converted into entities and nested facts are converted into triples (see Appendix for details).
For our approach, we implement three variants of NestE: NestE-Q (using quaternions), NestE-H (using hyperbolic quaternions), NestE-S (split quaternions), as well as their counterparts with translations: NestE-QB, NestE-HB, and NestE-SB. 
In the Appendix, we also extend BiVE-Q and BiVE-B to other hypercomplex numbers: BiVE-H, BiVE-HB BiVE-S, and BiVE-SB for further comparison. 
We employ three standard metrics: Filtered MR (Mean Rank), MRR (Mean Reciprocal Rank), and Hit@$10$. 
We report the mean performance over $10$ random seeds for each method, and the relatively small standard deviations are omitted. 


\paragraph{Implementation details} 
We implement the framework based on OpenKE \footnote{https://github.com/thunlp/OpenKE} and the code  \footnote{https://github.com/bdi-lab/BiVE/}.
We train our methods on triple prediction and evaluate them on other tasks. The dimensionality is set to be $d=200$. 
We reuse the hyperparameters as used in \cite{DBLP:conf/aaai/Chanyoung} and do not perform further hyperparameter searches. The detailed hyperparameter settings can be found in the Appendix.

\begin{table*}[t!]
\small
\centering
\setlength{\tabcolsep}{0.3em}
\renewcommand{\arraystretch}{0.9} 
\caption{Results of triple prediction. Shaded numbers are better results than the best baseline. The best scores are boldfaced and the second best scores are underlined. * denotes results taking from \cite{DBLP:conf/aaai/Chanyoung}. }
\vspace{-0.2cm}
\begin{tabular}{cccccccccc}
\toprule
 & \multicolumn{3}{c}{\fbh} & \multicolumn{3}{c}{\fbhe} & \multicolumn{3}{c}{\dbhe} \\
 & MR ($\downarrow$) & MRR  ($\uparrow$) & Hit@10 ($\uparrow$) & MR ($\downarrow$) & MRR ($\uparrow$) & Hit@10 ($\uparrow$) & MR ($\downarrow$) & MRR ($\uparrow$) & Hit@10 ($\uparrow$) \\
\midrule
QuatE*  & 145603.8 & 0.103 & 0.114 & 94684.4 & 0.101 & 0.209 & 26485.0 & 0.157 & 0.179 \\
BiQUE*  & 81687.5 & 0.104 & 0.115 & 61015.2 & 0.135 & 0.205 & 19079.4 & 0.163 & 0.185 \\
\midrule
Neural-LP* & 115016.6 &  0.070 &  0.073 &  90000.4 &  0.238 &  0.274 &  21130.5 &  0.170 & 0.209 \\
DRUM* & 115016.6 & 0.069 & 0.073 & 90000.3 & 0.261 &  0.274 & 21130.5 & 0.166 & 0.209 \\
AnyBURL* & 108079.6 & 0.096 & 0.108 & 83136.8 & 0.191 & 0.252 & 20530.8 & 0.177 & 0.214 \\
\midrule
BiVE-Q & 6.20 & 0.855 &	0.941 & 8.35 & 0.711 & 0.866 & 3.63 & 0.687 & 0.958 \\
BiVE-B & 8.63 & 0.833 & 0.924 & 9.53 & 0.705 & 0.860 & 4.66 & 0.718 & 0.945\\
\midrule
NestE-Q (Ours) & 6.56 & \cellcolor{gray!25}0.863 & \cellcolor{gray!25}0.953 & \cellcolor{gray!25}5.77 & \cellcolor{gray!25}0.811 & \cellcolor{gray!25}0.943 & \cellcolor{gray!25}3.51 & \cellcolor{gray!25}0.809 & \cellcolor{gray!25}0.960  \\
NestE-H (Ours) & \cellcolor{gray!25}4.69 & \cellcolor{gray!25}0.858 & \cellcolor{gray!25}0.964 & \cellcolor{gray!25}3.99 & \cellcolor{gray!25}0.781 & \cellcolor{gray!25}0.943 & \cellcolor{gray!25}2.65 & \cellcolor{gray!25}0.806 & \cellcolor{gray!25}0.969 \\
NestE-S (Ours) & \cellcolor{gray!25}\underline{3.87} & \cellcolor{gray!25}0.867 & \cellcolor{gray!25}\underline{0.977} & \cellcolor{gray!25}3.60 & \cellcolor{gray!25}0.795 & \cellcolor{gray!25}0.947 & \cellcolor{gray!25}2.55 & \cellcolor{gray!25}0.809 & \cellcolor{gray!25}0.966  \\
\midrule
NestE-QB (Ours) & \cellcolor{gray!25}6.04 & \cellcolor{gray!25}0.898 & \cellcolor{gray!25}0.958 & \cellcolor{gray!25}5.55 & \cellcolor{gray!25}\underline{0.845} & \cellcolor{gray!25}0.947 & \cellcolor{gray!25}\underline{2.54} & \cellcolor{gray!25}\underline{0.847} & \cellcolor{gray!25}\underline{0.973}  \\
NestE-HB (Ours) & \cellcolor{gray!25}3.82 & \cellcolor{gray!25}\underline{0.899} & \cellcolor{gray!25}0.971 & \cellcolor{gray!25}\underline{3.53} &  \cellcolor{gray!25}0.828 & \cellcolor{gray!25}\underline{0.955} & \cellcolor{gray!25}2.62 & \cellcolor{gray!25}0.842 & \cellcolor{gray!25}0.972 \\
NestE-SB (Ours) & \cellcolor{gray!25}\textbf{3.34} & \cellcolor{gray!25}\textbf{0.922} & \cellcolor{gray!25}\textbf{0.982} & \cellcolor{gray!25}\textbf{3.05} & \cellcolor{gray!25}\textbf{0.851} & \cellcolor{gray!25}\textbf{0.962} & \cellcolor{gray!25}\textbf{2.07} & \cellcolor{gray!25}\textbf{0.862} & \cellcolor{gray!25}\textbf{0.984} \\
\bottomrule
\end{tabular}
\vspace{-0.1cm}
\label{tb:tp}
\end{table*}

\begin{table*}[t!]
\small
\centering
\setlength{\tabcolsep}{0.3em}
\renewcommand{\arraystretch}{0.9} 
\caption{Results of conditional link prediction. Shaded numbers are better results than the best baseline. The best scores are boldfaced and the second best scores are underlined. * denotes results taking from \cite{DBLP:conf/aaai/Chanyoung}. }
\vspace{-0.2cm}
\begin{tabular}{cccccccccc}
\toprule
 & \multicolumn{3}{c}{\fbh} & \multicolumn{3}{c}{\fbhe} & \multicolumn{3}{c}{\dbhe} \\
 & MR ($\downarrow$) & MRR ($\uparrow$) & Hit@10 ($\uparrow$) & MR ($\downarrow$) & MRR ($\uparrow$) & Hit@10 ($\uparrow$) & MR ($\downarrow$) & MRR ($\uparrow$) & Hit@10 ($\uparrow$) \\
\midrule
QuatE* & 163.7 & 0.346 & 0.494 & 1546.4 & 0.124 & 0.189 & 551.6 & 0.208 & 0.309 \\
BiQUE* & 111.0 & 0.423 & 0.641 & 90.1 & 0.387 & 0.617 & 29.5 & 0.378 & 0.677 \\
\midrule
Neural-LP* & 185.9 & 0.433 & 0.648 & 146.2 & 0.466 & 0.716 & 32.2 & 0.517 & 0.756\\
DRUM* & 262.7 & 0.394 & 0.555 & 207.6 & 0.413 & 0.620 & 49.0 & 0.470 & 0.732\\
AnyBURL* & 228.5 & 0.380 & 0.563 & 166.0 & 0.418 & 0.607 & 81.7 & 0.403 & 0.594\\
\midrule
BiVE-Q & 4.33 & 0.826 & 0.948 & 6.56 & 0.761 & 0.886 & 2.69 & 0.852 & 0.971  \\
BiVE-B & 5.34 & 0.836 & 0.940 & 7.49 & 0.761 & 0.872 & 2.91 & 0.858 & 0.967  \\
\midrule
NestE-Q (Ours)  & \cellcolor{gray!25}1.70 & \cellcolor{gray!25}0.930 & \cellcolor{gray!25}0.986 & \cellcolor{gray!25}2.89 & \cellcolor{gray!25}0.863 & \cellcolor{gray!25}0.948 & \cellcolor{gray!25}\textbf{1.68} & \cellcolor{gray!25}\underline{0.930} & \cellcolor{gray!25}0.987  \\

NestE-H (Ours) & \cellcolor{gray!25}1.68 & \cellcolor{gray!25}0.909 &	\cellcolor{gray!25}0.987 & \cellcolor{gray!25}2.87 & \cellcolor{gray!25}0.843 & \cellcolor{gray!25}0.945 & \cellcolor{gray!25}1.82 & \cellcolor{gray!25}0.912 & \cellcolor{gray!25}0.986 \\
NestE-S (Ours)  & \cellcolor{gray!25}\underline{1.54} & \cellcolor{gray!25}0.925 & \cellcolor{gray!25}\textbf{0.991} & \cellcolor{gray!25}3.04 & \cellcolor{gray!25}0.850 & \cellcolor{gray!25}0.941 & \cellcolor{gray!25}1.76 & \cellcolor{gray!25}0.910 & \cellcolor{gray!25}\underline{0.988}  \\
\midrule
NestE-QB (Ours)  & \cellcolor{gray!25}1.71 & \cellcolor{gray!25}\textbf{0.935} & \cellcolor{gray!25}0.987 & \cellcolor{gray!25}3.00 & \cellcolor{gray!25}\underline{0.865} & \cellcolor{gray!25}0.949 & \cellcolor{gray!25}\underline{1.70} & \cellcolor{gray!25}\textbf{0.931} & \cellcolor{gray!25}0.986  \\
Fact-HB (Ours) & \cellcolor{gray!25}1.60 & \cellcolor{gray!25}0.924 & \cellcolor{gray!25}\underline{0.989} & \cellcolor{gray!25}\underline{2.76} & \cellcolor{gray!25}0.855 & \cellcolor{gray!25}\underline{0.950} & \cellcolor{gray!25}1.92 & \cellcolor{gray!25}0.918 & \cellcolor{gray!25}0.981 \\
NestE-SB (Ours) & \cellcolor{gray!25}\textbf{1.52} & \cellcolor{gray!25}\underline{0.934} & \cellcolor{gray!25}\textbf{0.991} & \cellcolor{gray!25}\textbf{2.61} & \cellcolor{gray!25}\textbf{0.867} & \cellcolor{gray!25}\textbf{0.951} & \cellcolor{gray!25}1.72 & \cellcolor{gray!25}0.919 & \cellcolor{gray!25}\textbf{0.990}  \\
\bottomrule
\end{tabular}
\vspace{-0.2cm}
\label{tb:clp}
\vspace{-0.2cm}
\end{table*}

\subsection{Main Results}

\paragraph{Triple prediction} 

Table \ref{tb:tp} presents the results of triple prediction. 
First, it shows that all triple-based approaches yield relatively modest results compared to BiVE-Q and BiVE-B, designed specifically for KGs with nested facts. Our approach, NestE-Q, the quaternionic version, already outperforms the baselines across most metrics. 
Particularly notable are the pronounced enhancements in FBHE and DBHE, with MRR improvements of 14.1\% and 17.7\% respectively, underscoring the efficacy of the proposed NestE model.
Furthermore, NestE-H and NestE-S demonstrate heightened performance over NestE-Q across various evaluation metrics, particularly in terms of MR. This highlights the advantages that hyperbolic quaternions and split quaternions offer over standard quaternions. Impressively, the split quaternionic version attains the highest performance, followed closely by the hyperbolic quaternionic variant.
Moreover, through the incorporation of a hypercomplex translation component, NestE-QB, Fact-HB, and NestE-SB consistently outperform their non-translation counterparts, illustrating the advantages of combining multiple transformations (rotation and translation) within the hypercomplex space.

\paragraph{Conditional link prediction} 
Table \ref{tb:clp} shows the outcomes of conditional link prediction. It is evident that all three NestE variants substantially outperform the two SoTA baselines, BiVE-Q and BiVE-B, across all datasets. Notably, the best NestE variant surpasses the baselines by 11.8\%, 13.9\%, and 8.5\% in terms of MRR for FBH, FBHE, and DBHE, respectively. This remarkable performance gain underscores the effectiveness of the proposed method.
Similar to the trends observed in triple prediction, the incorporation of translation components in NestE-QB, Fact-HB, and Fact-SB leads to further improvements over their counterparts without translation components. This reaffirms the advantages gained from the integration of multiple hypercomplex transformations. Intriguingly, we noticed that varying hypercomplex number systems yield the best performance on different datasets, contrasting the observations from triple prediction. We conjecture that this stems from the inherent variance in inductive biases offered by different hypercomplex number systems, making them more suitable for certain datasets over others.
We believe the choices of spaces can be linked to a hyperparameter that offers flexibility in adapting to diverse dataset characteristics.

\begin{table}
\small
\centering
\setlength{\tabcolsep}{0.2em}
\renewcommand{\arraystretch}{0.8} 
\caption{Results of base link prediction. The best scores are boldfaced and the second best scores are underlined. * denotes results taking from \cite{DBLP:conf/aaai/Chanyoung}. }
\vspace{-0.3cm}
\resizebox{\columnwidth}{!}
{
\begin{tabular}{ccccccc}
\toprule
 & \multicolumn{3}{c}{\fbhe} & \multicolumn{3}{c}{\dbhe} \\
 & MR ($\downarrow$) & MRR ($\uparrow$) & Hit@10 ($\uparrow$) & MR ($\downarrow$) & MRR ($\uparrow$) & Hit@10 ($\uparrow$) \\

\midrule
QuatE* & 139.0 & 0.354 & 0.581 & 409.6 & 0.264 & 0.440\\
BiQUE* &  134.9 & 0.356 & 0.583 & \textbf{376.6} & 0.274 & \textbf{0.446}\\
\midrule
Neural-LP* & 1942.5 & 0.315 & 0.486 & 2904.8 & 0.233 & 0.357 \\ 
DRUM* & 1945.6 & 0.317 & 0.490 & 2904.7 & 0.237 & 0.359 \\
AnyBURL* & 342.0 & 0.310 & 0.526 & 879.1 & 0.220 & 0.364\\
\midrule
BiVE-Q &  136.13 & 0.369 & 0.603 & 827.18 & 0.271 & 0.428  \\
BiVE-B & 136.54 & \underline{0.370} & \underline{0.607} & 795.59 & 0.274 & 0.422 \\
\midrule
NestE-Q (Ours) &  \underline{131.72} & 0.365 & 0.605  & 749.75 & \underline{0.284} & \textbf{0.446}  \\
Fact-H (Ours) & 153.00 & 0.349 & 0.593 & 868.82 & 0.266 & 0.423  \\
NestE-S (Ours) & 149.64 & 0.350 & 0.592 & 895.85 & 0.272 & 0.432   \\
\midrule
NestE-QB (Ours) & \textbf{130.13} & \textbf{0.371} & \textbf{0.608} & 751.18 & \textbf{0.289} & \underline{0.443}  \\
Fact-HB (Ours) & 155.74 & 0.353 & 0.594 & 801.76 & 0.271 & 0.423 \\
NestE-SB (Ours) & 149.73 & 0.355 & 0.594 & 827.89 & 0.273 & 0.431  \\
\bottomrule
\end{tabular}
}
\label{tb:blp}
\end{table}

\paragraph{Base link prediction}
Table \ref{tb:blp} illustrates the results of base link prediction. Among our approaches, namely NestE-Q, NestE-H, and NestE-S, we observe competitive or improved results in comparison to SoTA embedding-based and rule-based methods on the FBHE and DBHE datasets. The best performance is achieved by NestE-QB, which outperforms the baselines across a majority of metrics. This outcome substantiates the fact that the incorporation of nested facts into triple-based KGs indeed enhances the inference capabilities for base link prediction.

\subsection{Ablation Analysis}

\begin{figure}
    \centering
    \includegraphics[width=\linewidth]{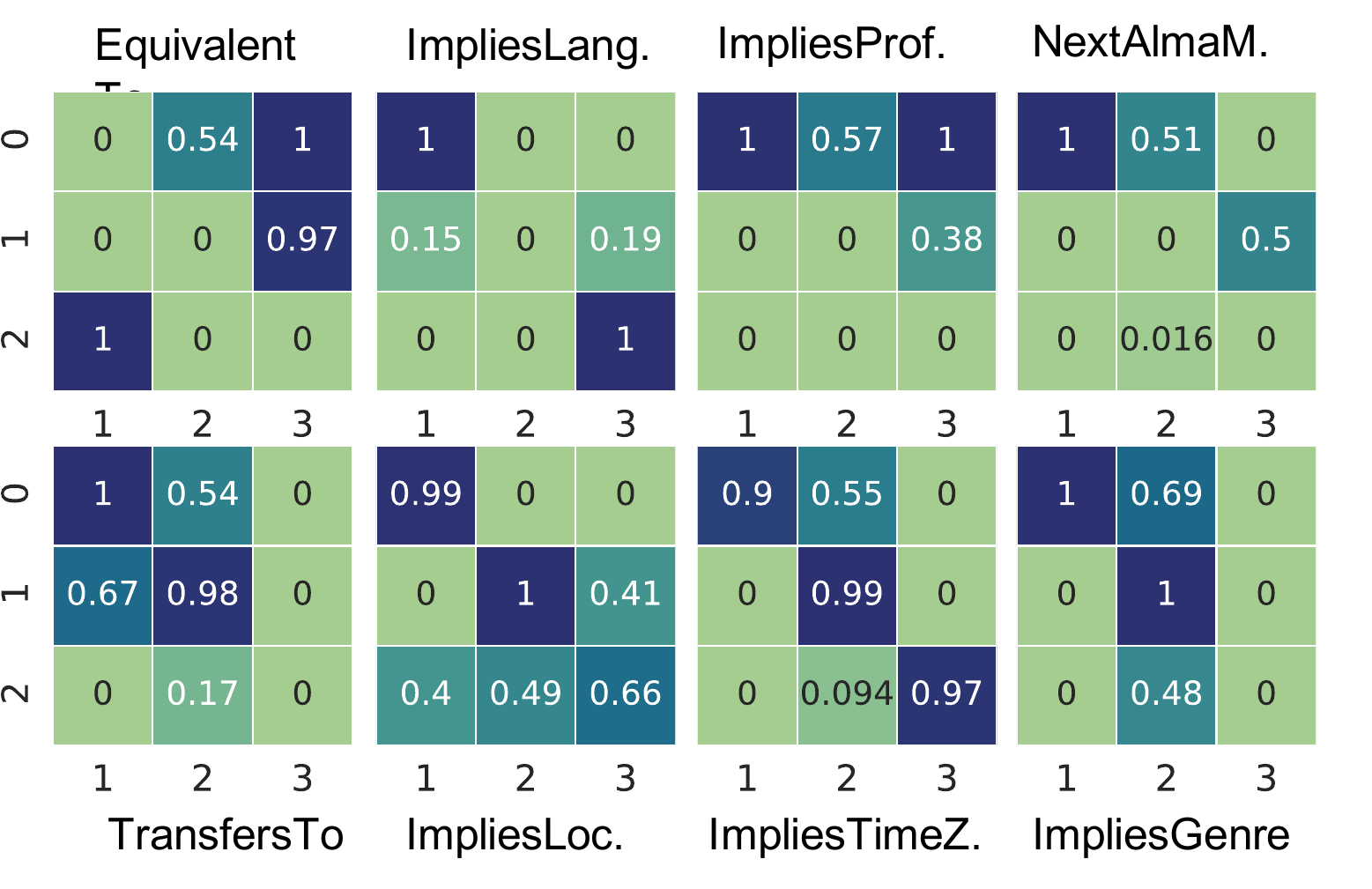}
     \vspace{-0.7cm}
    \caption{The visualization of the average of the real component embeddings of the $8$ nested relations in DBHE. }
    \vspace{-0.3cm}
    \label{fig:patternanalysis}
\end{figure}

\paragraph{Embedding analysis of logical patterns} 
To verify whether the learned embeddings capture the inference of logical patterns over nested facts, 
we visualized the real part of the embeddings of the $8$ relations in DBHE. 
The analysis of the embeddings yields insightful observations. As shown in Fig. \ref{fig:patternanalysis}, the lower left element and upper right element of the embedding \emph{EquivalentTo} are $1$, showcasing that \emph{EquivalentTo} predominantly adheres to R-symmetry or R-inverse. On the other hand, the upper left element and lower right element of the \emph{ImpliesLang.} are $1$, affirming its alignment with the R-implication rule. Similarly, the embeddings of \emph{NextAlmaM.}, \emph{TransfersTo}, and \emph{ImpliesGenre} indicate high adherence to E-implicationsas as only one of the corners is $1$. 
We find that the embedding of relation \emph{ImpliesProf.} does not have a significant pattern. We conjecture that this is because \emph{ImpliesProf.} follows many rule patterns and there exists no global solution that satisfies all rules. 
See the Appendix for the statistics of the logical patterns in the datasets.

\paragraph{Influence of nested fact embeddings}
To evaluate the influence of nested fact embeddings, we perform a comparison by excluding the loss associated with nested fact embeddings (i.e., setting $\lambda_1=0$). The outcomes presented in Table \ref{tb:impact_nested} underscore the significant enhancements achieved by incorporating nested fact embeddings, particularly evident in the improvements in MR and H@10 for DBHE.

\begin{table}
\small
\centering
\setlength{\tabcolsep}{0.2em}
\renewcommand{\arraystretch}{0.9} 
\caption{Ablation study on the nested fact embeddings for base link prediction. 
Best results are boldfaced. 
}
\vspace{-0.3cm}
\resizebox{\columnwidth}{!}
{
\begin{tabular}{ccccccc}
\toprule
 & \multicolumn{3}{c}{\fbhe} & \multicolumn{3}{c}{\dbhe} \\
 & MR ($\downarrow$) & MRR ($\uparrow$) & Hit@10 ($\uparrow$) & MR ($\downarrow$) & MRR ($\uparrow$) & Hit@10 ($\uparrow$) \\
\midrule
NestE-Q ($\lambda_1=0$) &  135.38 & \textbf{0.368} & 0.604 & 799.97 & 0.281 & 0.431  \\
NestE-Q ($\lambda_1=0.5$) &  \textbf{131.72} & 0.365 & \textbf{0.605}  & \textbf{749.75} & \textbf{0.284} & \textbf{0.446}   \\
\midrule
NestE-H ($\lambda_1=0$) &  154.75 &  0.347 & 0.589 & 922.34 & 0.267 & 0.420 \\
NestE-H ($\lambda_1=0.5$) & \textbf{153.00} & \textbf{0.349} & \textbf{0.593} & \textbf{868.82} & 0.266 & \textbf{0.423}  \\
\midrule
NestE-S ($\lambda_1=0$) & 151.84 & 0.347 & 0.589 & 910.16 & \textbf{0.272} & 0.427    \\
NestE-S ($\lambda_1=0.5$)& \textbf{149.64} & \textbf{0.350} & \textbf{0.592} & \textbf{895.85} & \textbf{0.272} & \textbf{0.432}  \\
\bottomrule
\end{tabular}
}
\vspace{-0.1cm}
\label{tb:impact_nested}
\end{table}

\begin{table}[t]
\scriptsize
\centering
\setlength{\tabcolsep}{0.2em}
\renewcommand{\arraystretch}{0.9} 
\caption{Performance per relation on triple prediction. Freq. indicates the number of nested facts in the test set.  }
\vspace{-0.3cm}
\resizebox{\columnwidth}{!}
{
\begin{tabular}{cccccccc}
\toprule
 $\rhat$ & Freq. & NestE-Q & NestE-QB & NestE-H & NestE-HB  & NestE-S & NestE-SB  \\
\midrule
Equiv.To & 98 & 0.994 & \textbf{0.997} & 0.992 & \textbf{0.997} & \textbf{0.997} & \underline{0.995} \\
ImpliesLang. & 29 & \underline{0.671} & 0.602 & \textbf{0.680} & 0.662 & 0.614 & 0.622 \\
ImpliesProf. & 210 & 0.807 & 0.916 & 0.830 & \underline{0.935} & 0.832 & \textbf{0.936} \\
ImpliesLocat. & 163 & \underline{0.929} & 0.869 & 0.893 & 0.810 & \underline{0.929} & \textbf{0.958} \\
ImpliesTime. & 44 & 0.305 & 0.297 & \underline{0.307} & \textbf{0.329} & 0.290 & 0.293 \\
ImpliesGenre & 84 & 0.726 & \underline{0.762} & 0.719 & 0.741 & 0.742 & \textbf{0.796} \\
NextAlmaM. & 14 & 0.770 & \textbf{0.812} & 0.689 & 0.688 & 0.751 & \underline{0.795} \\
Transf.To & 29 &  \textbf{0.977} & \underline{0.952} & 0.964 & 0.949 & 0.921 & 0.953 \\
\bottomrule
\end{tabular}
}

\label{tb:rel}
\vspace{-0.5cm}
\end{table}

\paragraph{Relation-specific performance}
In Table \ref{tb:rel}, we present the performance results for each relation within the DBHE dataset. Notably, the diverse hypercomplex number systems lead to optimal performance for different relations. This reiterates our conjecture that distinct benefits are offered by varying hypercomplex number systems, catering to the specific characteristics of different relation types.
Remarkably, our findings reveal that the incorporation of a hypercomplex translation component (as seen in NestE-QB, NestE-HB, and NestE-SB) notably enhances the embeddings of relations such as \emph{ImpliesProf.} and \emph{ImpliesGenre} across all variants of hypercomplex number systems. 
However, this does not extend to relations like \emph{ImpliesLocat.} and \emph{ImpliesLang.}, suggesting a more complex relationship between these specific relations and the hypercomplex translation.

\section{Conclusion}

This paper considers a novel perspective by extending traditional atomic factual knowledge representation to include nested factual knowledge. This enables the representation of both temporal situations and logical patterns that go beyond conventional first-order logic expressions (Horn rules). Our proposed approach, NestE, presents a family of hypercomplex embeddings capable of embedding both atomic and nested factual knowledge. This framework effectively captures essential logical patterns that emerge from nested facts. Empirical evaluation demonstrates the substantial performance enhancements achieved by NestE compared to existing baseline methods. 
Additionally, our generalized hypercomplex embedding framework unifies previous algebraic (e.g., quaternionic) and geometric (e.g., hyperbolic) embedding methods, offering versatility in embedding diverse relation types. 
One under-explored version of our method is to consider scoring function based on distance instead of inner product.  
Another interesting avenue for future exploration involves devising efficient strategies for selecting appropriate algebraic and geometric operations tailored to specific relations in KGs.

\section*{Acknowledgement}

The authors thank the International Max Planck Research School for Intelligent Systems (IMPRS-IS) for supporting Bo Xiong and Zihao Wang. 
This research has been partially funded by Deutsche Forschungsgemeinschaft (DFG, German Research Foundation) under Germany’s Excellence Strategy - EXC 2075 - 390740016. We acknowledge the support by the Stuttgart Center for Simulation Science (SimTech).
Bo Xiong is funded by the European Union’s Horizon 2020 research and innovation programme under the Marie Skłodowska-Curie grant agreement No: 860801.
Zihao Wang was funded by the the Bundesministerium für Wirtschaft und Energie (BMWi), grant aggrement No. 01MK20008F.

\bibliography{main}

\newpage
\newpage

\appendix

\onecolumn
\section*{\Large{Appendix}}

\section{Supplementary Related Work}

\paragraph{Triple-shaped Representations}

The majority of KG embedding models focus on triple-shaped KGs, where each fact is represented as a triple. Notable instances include the \emph{additive} (or \emph{translational}) family \cite{DBLP:conf/nips/BordesUGWY13, DBLP:conf/aaai/WangZFC14, DBLP:conf/iclr/SunDNT19}. This family embodies facts as translations, with representations like $\mathbf{h}+\mathbf{r} \approx \mathbf{t}$ or rotations as $\mathbf{h} \circ \mathbf{r} \approx \mathbf{t}$. 
On the other hand, the \emph{multiplicative} (or \emph{semantic matching}) family \cite{DBLP:conf/icml/NickelTK11, DBLP:journals/corr/YangYHGD14a}, models the relationship between two entities through bilinear interactions of $\mathbf{h}$, $\mathbf{r}$, and $\mathbf{t}$. 
The translational family excels in inferring various inference patterns (e.g., symmetry, inversion, and composition), while the multiplicative family facilitates expressive relational mapping (e.g., one-to-many and many-to-many) between head and tail entities. Notably, certain approaches \cite{DBLP:conf/icml/TrouillonWRGB16, DBLP:conf/nips/0007TYL19} amalgamate the strengths of both families. 
A prime example, ComplEx \cite{DBLP:conf/icml/TrouillonWRGB16}, embeds entities and relations in a complex space and employs the Hermitian product, possessing geometric rotation significance, to model diverse patterns. 
QuatE \cite{DBLP:conf/nips/0007TYL19} and BiQUE \cite{DBLP:conf/emnlp/GuoK21} further extends ComplEx to a 4-dimensional hypercomplex space with quaternion representations. This extension empowers rich and expressive mapping between head and tail entities, facilitated by relational rotation endowed with additional degrees of freedom. Other methods \cite{DBLP:conf/acl/ChamiWJSRR20,DBLP:conf/nips/BalazevicAH19,DBLP:conf/acl/NayyeriXMAA0S23} embed entities in hyperbolic space, which excels in modeling hierarchical structures. 
Our hypercomplex embedding method unifies these algebraic and geometric embeddings in a single framework.


\section{Theoretical Justifications}

We extend the vanilla logical patterns in KGs to include nested facts. 
This can be expressed in a non-first-order-logic-like form $\psi \stackrel{\widehat{r}}{\rightarrow}\phi$.

\begin{itemize}
    \item \textbf{Relational symmetry (R-symmetry):} an atomic relation $r$ is symmetric w.r.t a nested relation $\widehat{r}$ if $\forall x,y \in \mathcal{E},  \langle x, r, y\rangle  \stackrel{\widehat{r}}{\leftrightarrow} \langle y, r, x\rangle$. 

    \item \textbf{Relational inverse (R-inverse):} two atomic relations $r_1$ and $r_2$ are inverse w.r.t a nested relation $\widehat{r}$ if $\forall x,y \in \mathcal{E}, $ $(\langle x, r_1, y\rangle \stackrel{\widehat{r}}{\leftrightarrow} \langle y, r_2, x\rangle)$.

    \item \textbf{Relational implication (R-implication):} an atomic relation $r_1$ implies a atomic relation $r_2$ w.r.t a nested relation $\widehat{r}$ if $\forall x,y \in \mathcal{E}, (\langle x, r_1, y\rangle \stackrel{\widehat{r}}{\rightarrow} \langle x, r_2, y\rangle)$.

    \item \textbf{Relational inverse implication (R-Inv-implication):} an atomic relation $r_1$ inversely implies an atomic relation $r_2$ w.r.t a nested relation $\widehat{r}$ if $\forall x,y \in \mathcal{E}, (\langle x, r_1, y\rangle \stackrel{\widehat{r}}{\rightarrow} \langle y, r_2, x\rangle)$.

    \item \textbf{Entity implication (E-implication):} an entity $x_1$ (resp. $y_1$) implies entity $x_2$ (resp. $y_2$) w.r.t an atomic relation $r$ and a nested relation $\widehat{r}$ if  $\forall y \in \mathcal{E}, (\langle x_1, r, y\rangle \stackrel{\widehat{r}}{\rightarrow} \langle x_2, r, y\rangle)$ (resp. $\forall x \in \mathcal{E}, (\langle x, r, y_1\rangle \stackrel{\widehat{r}}{\rightarrow} \langle x, r, y_2\rangle)$ ).

    \item \textbf{Entity relational implication (E-R-implication):} an entity $x_1$ and relation $r_1$ (resp. $y_1$ and relation $r_1$) implies entity $x_2$ and relation $r_2$ (resp. $y_2$ and relation $r_2$) 
    w.r.t a nested relation $\widehat{r}$ if  $\forall y \in \mathcal{E}, (\langle x_1, r_1, y\rangle \stackrel{\widehat{r}}{\rightarrow} \langle x_2, r_2, y\rangle)$ (resp. $\forall x \in \mathcal{E}, (\langle x, r_1, y_1\rangle \stackrel{\widehat{r}}{\rightarrow} \langle x, r_2, y_2\rangle)$).

    \item \textbf{Entity relational inverse implication (E-R-Inv-implication):}  an entity $x_1$ and relation $r_1$ (resp. $y_1$ and relation $r_1$) inversely implies entity $x_2$ and relation $r_2$ (resp. $y_2$ and relation $r_2$) w.r.t a nested relation $\widehat{r}$ if  $\forall y \in \mathcal{E}, (\langle x_1, r_1, y\rangle \stackrel{\widehat{r}}{\rightarrow} \langle y, r_2, x_2\rangle)$ (resp. $\forall x \in \mathcal{E}, (\langle x, r_1, y_1\rangle \stackrel{\widehat{r}}{\rightarrow} \langle y_2, r_2, x\rangle)$).
     \item \textbf{Dual Entity implication (Dual E-implication):} an entity pair ($x_1$, $x_2$) implies another entity pair ($y_1$, $y_2$)  iff both ($x_1$, $y_1$) and ($x_2$, $y_2$) satisfy E-implication. 

\end{itemize}

\begin{proposition}
\label{prop:NestE}
    NestE can infer R-symmetry, R-inverse, R-implication, R-Inv-implication, E-implication, E-R-implication, E-R-Inv-implication, and Dual E-implication.
\end{proposition}

To infer different logical patterns via different free variables, we can set some elements of the relation matrix to be zero-valued or one-valued complex numbers. 
We prove proposition \ref{prop:NestE} by showing the following special solutions for each case of pattern. 

\begin{proposition}
    Relational symmetry can be inferred by setting
    \begin{equation}
        \mathbf{R} = 
         \left[ 
              {\begin{array}{ccc}
               \mathbf{0} & \mathbf{0} & \mathbf{1}\\
               \mathbf{0} & \mathbf{R}_{22} & \mathbf{0}\\
               \mathbf{1} & \mathbf{0} & \mathbf{0}\\
              \end{array} } 
        \right]
    \end{equation}  
    where $\mathbf{r} \otimes \mathbf{R}_{22} = \mathbf{r}$ implying $\mathbf{R}_{22}=\mathbf{1}$. 
\end{proposition}

\begin{proposition}
    Relational inversion can be inferred by setting
    \begin{equation}
        \mathbf{R} = 
         \left[ 
              {\begin{array}{ccc}
               \mathbf{0} & \mathbf{0} & \mathbf{1}\\
               \mathbf{0} & \mathbf{R}_{22} & \mathbf{0}\\
               \mathbf{1} & \mathbf{0} & \mathbf{0}\\
              \end{array} } 
        \right]
    \end{equation}  
    where $\mathbf{r}_1 \otimes \mathbf{R}_{22} = \mathbf{r}_2$ and $\mathbf{r}_2 \otimes \mathbf{R}_{22} = \mathbf{r}_1$ implying $\mathbf{R}_{22} = \mp 1$.

\end{proposition}

\begin{proposition}
    Relational implication can be inferred by setting
    \begin{equation}
        \mathbf{R} = 
         \left[ 
              {\begin{array}{ccc}
               \mathbf{1} & \mathbf{0} & \mathbf{0}\\
               \mathbf{0} & \mathbf{R}_{22} & \mathbf{0}\\
               \mathbf{0} & \mathbf{0} & \mathbf{1}\\
              \end{array} } 
        \right]
    \end{equation}  
    where $\mathbf{r}_1 \otimes \mathbf{R}_{22} = \mathbf{r}_2$.
    
\end{proposition}

\begin{proposition}
    Relational inverse implication can be inferred by setting
    \begin{equation}
        \mathbf{R} = 
         \left[ 
              {\begin{array}{ccc}
               \mathbf{0} & \mathbf{0} & \mathbf{1}\\
               \mathbf{0} & \mathbf{R}_{22} & \mathbf{0}\\
               \mathbf{1} & \mathbf{0} & \mathbf{0}\\
              \end{array} } 
        \right]
    \end{equation}  
    where $\mathbf{r}_1 \otimes \mathbf{R}_{22} = \mathbf{r}_2$.
\end{proposition}

\begin{proposition}
    Entity implication can be inferred by setting
    \begin{equation}
        \mathbf{R} = 
         \left[ 
              {\begin{array}{ccc}
               \mathbf{R}_{11} & \mathbf{R}_{12} & \mathbf{0}\\
               \mathbf{R}_{21} & \mathbf{R}_{21} & \mathbf{0}\\
               \mathbf{0} & \mathbf{0} & \mathbf{1}\\
              \end{array} } 
        \right]
    \end{equation}  
    where   
    \begin{align}
    \mathbf{x}_1 \otimes \mathbf{R}_{11} + \mathbf{r} \otimes  \mathbf{R}_{21} = \mathbf{x}_2 \\
    \mathbf{x}_1 \otimes \mathbf{R}_{12} + \mathbf{r} \otimes  \mathbf{R}_{22} = \mathbf{r}
      \end{align}
    or 
    
    \begin{equation}
        \mathbf{R} = 
         \left[ 
              {\begin{array}{ccc}
               \mathbf{1} & \mathbf{0} & \mathbf{0}\\
               \mathbf{0} & \mathbf{R}_{22} & \mathbf{R}_{23}\\
               \mathbf{0} & \mathbf{R}_{32} & \mathbf{R}_{33}\\
              \end{array} } 
        \right]
    \end{equation}  
    where 
    \begin{align}
        \mathbf{r} \otimes  \mathbf{R}_{32} + \mathbf{y}_1 \otimes  \mathbf{R}_{33} = \mathbf{y}_2\\
        \mathbf{r} \otimes  \mathbf{R}_{22} + \mathbf{y}_1 \otimes \mathbf{R}_{32}  = \mathbf{r}
    \end{align}
    
\end{proposition}

\begin{proposition}
    Entity inverse implication can be inferred by setting
    \begin{equation}
        \mathbf{R} = 
         \left[ 
              {\begin{array}{ccc}
               \mathbf{0} & \mathbf{0} & \mathbf{R}_{13}\\
               \mathbf{0} & \mathbf{1} & \mathbf{0}\\
               \mathbf{R}_{31} & \mathbf{0} & \mathbf{1}\\
              \end{array} } 
        \right]
    \end{equation}  
    where $\mathbf{x}_1 \otimes \mathbf{R}_{31} = \mathbf{x}_2$. Or,
    \begin{equation}
        \mathbf{R} = 
         \left[ 
              {\begin{array}{ccc}
               \mathbf{0} & \mathbf{0} & \mathbf{R}_{13}\\
               \mathbf{0} & \mathbf{1} & \mathbf{0}\\
               \mathbf{R}_{31} & \mathbf{0} & \mathbf{0}\\
              \end{array} } 
        \right]
    \end{equation}  
    where $\mathbf{y}_1 \otimes \mathbf{R}_{33} = \mathbf{y}_2$.
\end{proposition}

\section{Experimental Details}


\subsection{Dataset Details}

The datasets \fbh, \fbhe, and \dbhe are constructed as detailed in \cite{DBLP:conf/aaai/Chanyoung}.
Both \fbh and \fbhe are derived from FB15K237, sourced from Freebase \cite{DBLP:conf/sigmod/BollackerEPST08}, while DBHE is based on DB15K from DBpedia \cite{DBLP:conf/semweb/AuerBKLCI07}.
FBH contains only nested facts that can be inferred from triple facts, including relations such as \emph{prerequisite\_for} and \emph{implies\_position}. In contrast, FBHE and DBHE incorporate additional externally-sourced information gathered from Wikipedia articles, giving rise to relations like \emph{next\_almaMater} and \emph{transfers\_to}.
For instance, the (vice)presidents of the United States and politicians' alma mater information were extracted by crawling relevant Wikipedia articles.
The datasets were meticulously curated, involving six weeks of manual effort to define and incorporate the nested facts \cite{DBLP:conf/aaai/Chanyoung}. The detailed data augmentation process can be found in Section 4 of the same paper.

Table \ref{tb:hlt_full} provides an overview of all nested relations and their corresponding examples that constitute FBH, FBHE, and DBHE. Specifically, FBHE contains ten higher-level relations, while FBH contains the initial six higher-level relations.
Certain relations within FBHE, namely WorksFor, SucceededBy, TransfersTo, and HigherThan, as well as NextAlmaMater and TransfersTo within DBHE, necessitated externally-sourced information. In the case of WorksFor, for instance, the authors crawled Wikipedia articles to obtain insights into the affiliations of individuals. Similarly, for HigherThan in FBHE, recent rankings from sources such as Fortune 1000 and Times University Ranking were leveraged to construct the respective triples.
Further details regarding the types of nested triples utilized in the creation of FBH, FBHE, and DBHE can be found in Table \ref{tb:fbh_full} and Table \ref{tb:dbh_full}. Additionally, these tables provide statistics pertaining to the logical patterns discussed in the paper.

\begin{table}[]
    \centering
    \begin{tabular}{c|c}
    Hyperparameter & value \\
    \midrule 
    learning rate $\alpha $ & 0.1 \\
    regularization rate $\beta$ & 0.1 \\
        $\lambda_1$ & 0.5  \\
        $\lambda_2$ & 0.2 \\
        dimensionality $d$ & 200 \\
     \midrule
    \end{tabular}
    \caption{The hyperparameter settings used in all experiments across all datasets.}
    \label{tb:hyperparameters}
\end{table}

\subsection{Implementation Details}

All experiments were conducted on machines equipped with 4 Nvidia A100 GPUs, each with 40GB of VRAM. Our framework was implemented based on OpenKE\footnote{https://github.com/thunlp/OpenKE} and the provided code\footnote{https://github.com/bdi-lab/BiVE/}. The training was performed using triple prediction, and evaluations were carried out for the other two tasks. The embedding dimensionality was set to $d=200$.

Hyperparameters utilized in this work were reused from \cite{DBLP:conf/aaai/Chanyoung}, and no additional hyperparameter tuning was performed. Specifically, all hyperparameters are listed in Table \ref{tb:hyperparameters}. The validation process occurred every 50 epochs, up to a maximum of 500 epochs. The best epoch for evaluation was selected based on the validation results.

\subsection{Supplemental Experiments}

\paragraph{Comparison of different hypercomplex spaces for BiVEs} 
We also implemented variants of BiVE in different hypercomplex spaces and present the results in Table \ref{tb:bive}. It was observed that the hyperbolic and split quaternionic versions of BiVE, denoted as BiVE-H and BiVE-S respectively, did not demonstrate significant improvements over the performance of BiVE-Q, even when leveraging their respective translation-enhanced versions. This phenomenon could potentially be attributed to the concatenation operation employed by BiVE and the use of a multilayer perceptron (MLP) for approximation. These factors might disrupt the geometric properties intrinsic to hypercomplex numbers, thus diminishing the potential advantages these geometries offer.

\begin{table*}[t!]
\small
\centering
\setlength{\tabcolsep}{0.3em}
\renewcommand{\arraystretch}{0.9} 
\caption{The performance comparison of BiVEs implemented via different hypercomplex spaces on triple prediction. }
\vspace{-0.2cm}
\begin{tabular}{cccccccccc}
\toprule
 & \multicolumn{3}{c}{\fbh} & \multicolumn{3}{c}{\fbhe} & \multicolumn{3}{c}{\dbhe} \\
 & MR ($\downarrow$) & MRR  ($\uparrow$) & Hit@10 ($\uparrow$) & MR ($\downarrow$) & MRR ($\uparrow$) & Hit@10 ($\uparrow$) & MR ($\downarrow$) & MRR ($\uparrow$) & Hit@10 ($\uparrow$) \\
\midrule
BiVE-Q & 6.20 & 0.855 &	0.941 & 8.35 & 0.711 & 0.866 & 3.63 & 0.687 & 0.958 \\
BiVE-H (Ours) & 6.70 & 0.838 & 0.934 & \cellcolor{gray!25}8.16 & 0.687 & 0.850 & 4.27 & 0.648 & 0.935 \\
BiVE-S (Ours) & \cellcolor{gray!25}6.13 & 0.824 & 0.938 & \cellcolor{gray!25}8.28 & 0.677 & 0.850 & 4.85 & 0.645 & 0.926  \\
\midrule
BiVE-B & 8.63 & 0.833 & 0.924 & 9.53 & 0.705 & 0.860 & 4.66 & 0.718 & 0.945\\
BiVE-HB (Ours) & 7.65 & 0.803 & 0.919 & 10.54 & 0.675 &	0.834 &  6.40 &  0.703 & 0.922  \\
BiVE-SB (Ours) & 9.77 & 0.799 & 0.914 & 9.83 & 0.678 & 0.853 & 7.58 & 0.712 & 0.914  \\
\bottomrule
\end{tabular}
\vspace{-0.1cm}
\label{tb:bive}
\end{table*}

\begin{table*}[htbp]
\scriptsize
\centering
\setlength{\tabcolsep}{0.15em}
\begin{tabular}{ccll}
\toprule
& $\rhat$ & \multicolumn{1}{c}{$\langle T_i,\rhat,T_j \rangle$} & Description \\
\midrule
\multirow{20}{*}{\fbhe} & \multirow{2}{*}{PrerequisiteFor} & $T_i$: (BAFTA\_Award, Nominates, The\_King's\_Speech) & For The King's Speech to win BAFTA Award, BAFTA Award should nominate The \\
 & & $T_j$: (The\_King's\_Speech, Wins, BAFTA\_Award) & King's Speech. \\
\cdashline{2-4}
 & \multirow{2}{*}{EquivalentTo} & $T_i$: (Hillary\_Clinton, IsMarriedTo, Bill\_Clinton) & \multirow{2}{*}{The two triples indicate the same information.} \\
 & & $T_j$: (Bill\_Clinton, IsMarriedTo, Hillary\_Clinton) & \\
\cdashline{2-4}
 & \multirow{2}{*}{ImpliesLocation} & $T_i$: (Sweden, CapitalIsLocatedIn, Stockholm) & \multirow{2}{*}{`The capital of Sweden is Stockholm' implies `Sweden contains Stockholm'.} \\
 & & $T_j$: (Sweden, Contains, Stockholm) & \\
\cdashline{2-4}
 & \multirow{2}{*}{ImpliesProfession} & $T_i$: (Liam\_Neeson, ActsIn, Love\_Actually) & \multirow{2}{*}{`Liam Neeson acts in Love Actually' implies `Liam Neeson is an actor'.} \\
 & & $T_j$: (Liam\_Neeson, IsA, Actor) & \\
\cdashline{2-4}
 & \multirow{2}{*}{ImpliesSports} & $T_i$: (Boston\_Red\_Socks, HasPosition, Infield) & \multirow{2}{*}{`Boston Red Socks has an infield position' implies `Boston Red Socks plays baseball'.} \\
 & & $T_j$: (Boston\_Red\_Socks, Plays, Baseball) & \\
\cdashline{2-4}
 & \multirow{2}{*}{NextEventPlace} & $T_i$: (1932\_Summer\_Olympics, IsHeldIn, Los\_Angeles) & Summer Olympics in 1932 and 1936 were held in Los Angeles and Berlin, respectively. \\
 & & $T_j$: (1936\_Summer\_Olympics, IsHeldIn, Berlin) & 1936 Summer Olympics is the next event of 1932 Summer Olympics.\\
\cdashline{2-4}
 & \multirow{2}{*}{WorksFor} & $T_i$: (Joe\_Biden, HoldsPosition, Vice\_President) & \multirow{2}{*}{Joe Biden was a vice president when Barack Obama was a president of the United States.} \\
 & & $T_j$: (Barack\_Obama, HoldsPosition, President) & \\
\cdashline{2-4}
 & \multirow{2}{*}{SucceededBy} & $T_i$: (George\_W.\_Bush, HoldsPosition, President) & \multirow{2}{*}{President Barack Obama succeeded President George W. Bush.} \\
 & & $T_j$: (Barack\_Obama, HoldsPosition, President) & \\
\cdashline{2-4}
 & \multirow{2}{*}{TransfersTo} & $T_i$: (David\_Beckham, PlaysFor, Real\_Madrid\_CF) & \multirow{2}{*}{David Beckham transferred from Real Madrid CF to LA Galaxy.} \\
 & & $T_j$: (David\_Beckham, PlaysFor, LA\_Galaxy) & \\
\cdashline{2-4}
 & \multirow{2}{*}{HigherThan} & $T_i$: (Walmart, IsRankedIn, Fortune\_500) & \multirow{2}{*}{Walmart is ranked higher than Bank of America in Fortune 500.} \\
 & & $T_j$: (Bank\_of\_America, IsRankedIn, Fortune\_500) & \\
\cdashline{1-4}
\multirow{16}{*}{\dbhe} & \multirow{2}{*}{EquivalentTo} & $T_i$: (David\_Beckham, IsMarriedTo, Victoria\_Beckham) & \multirow{2}{*}{The two triples indicate the same information.} \\
 & & $T_j$: (Victoria\_Beckham, IsMarriedTo, David\_Beckham) & \\
\cdashline{2-4}
 & \multirow{2}{*}{ImpliesLanguage} & $T_i$: (Italy, HasOfficialLanguage, Italian\_Language) & `The official language of Italy is the Italian language' implies `The Italian language is \\
 & & $T_j$: (Italy, UsesLanguage, Italian\_Language) & used in Italy'. \\
\cdashline{2-4}
 & \multirow{2}{*}{ImpliesProfession} & $T_i$: (Psycho, IsDirectedBy, Alfred\_Hitchcock) & \multirow{2}{*}{`Psycho is directed by Alfred Hitchcock' implies `Alfred Hitchcock is a film producer'.} \\
 & & $T_j$: (Alfred\_Hitchcock, IsA, Film\_Producer) & \\
\cdashline{2-4}
 & \multirow{2}{*}{ImpliesLocation} & $T_i$: (Mariah\_Carey, LivesIn, New\_York\_City) & \multirow{2}{*}{`Mariah Carey lives in New York City' implies `Mariah Carey lives in New York'} \\
 & & $T_j$: (Mariah\_Carey, LivesIn, New\_York) & \\
\cdashline{2-4}
 & \multirow{2}{*}{ImpliesTimeZone} & $T_i$: (Czech\_Republic, TimeZone, Central\_European) & `Czech Republic is included in Central European Time Zone' implies `Prague is included \\
 & & $T_j$: (Prague, TimeZone, Central\_European) & in Central European Time Zone'. \\
\cdashline{2-4}
 & \multirow{2}{*}{ImpliesGenre} & $T_i$: (Pharrell\_Williams, Genre, Progressive\_Rock) & `Pharrell Williams is a progressive rock musician' implies `Pharrell Williams is a rock \\
 & & $T_j$: (Pharrell\_Williams, Genre, Rock\_Music) & musician'. \\
\cdashline{2-4}
 & \multirow{2}{*}{NextAlmaMater} & $T_i$: (Gerald\_Ford, StudiesIn, University\_of\_Michigan) & \multirow{2}{*}{Gerald Ford studied in University of Michigan. Then, he studied in Yale University.} \\
 & & $T_j$: (Gerald\_Ford, StudiesIn, Yale\_University) & \\
\cdashline{2-4}
 & \multirow{2}{*}{TransfersTo} & $T_i$: (Ronaldo, PlaysFor, FC\_Barcelona) & \multirow{2}{*}{Ronaldo transferred from FC Barcelona to Inter Millan.} \\
 & & $T_j$: (Ronaldo, PlaysFor, Inter\_Millan) & \\
\bottomrule
\end{tabular}
\caption{The nested factual relations and the corresponding examples of the nested triples used to create \fbhe, and \dbhe. Table taken from \cite{DBLP:conf/aaai/Chanyoung}.}
\label{tb:hlt_full}
\end{table*}

\begin{table*}[htbp]
\scriptsize
\centering
\setlength{\tabcolsep}{1em}
\begin{tabular}{cllc}
\toprule
 & & Example & Frequency \\
\midrule
\multirow{6}{*}{$\langle T_i$, PrerequisiteFor, $T_j \rangle$} & $T_i$: (Person A, DatesWith, Person B) & (Bruce\_Willis, DatesWith, Demi\_Moore) & \multirow{2}{*}{222} \\
 & $T_j$: (Person A, BreaksUpWith, Person B) & (Bruce\_Willis, BreaksUpWith, Demi\_Moore) & \\
\cdashline{2-4}
 & $T_i$: (Award A, Nominates, Work B) & (BAFTA\_Award, Nominates, The\_King's\_Speech) & \multirow{2}{*}{2,335} \\
 & $T_j$: (Work B, Wins, Award A) & (The\_King's\_Speech, Wins, BAFTA\_Award) & \\
\cdashline{2-4}
 & $T_i$: (Person A, HasNationality, Country B) & (Neymar, HasNationality, Brazil) & \multirow{2}{*}{109} \\
 & $T_j$: (Person A, PlaysFor, National Team of B) & (Neymar, PlaysFor, Brazil\_National\_Football\_Team) &  \\
\cdashline{1-4}
\multirow{8}{*}{$\langle T_i$, EquivalentTo, $T_j \rangle$} & $T_i$: (Person A, IsASiblingTo, Person B) & (Serena\_Williams, IsASiblingTo, Venus\_Williams) & \multirow{2}{*}{120} \\
 & $T_j$: (Person B, IsASiblingTo, Person A) & (Venus\_Williams, IsASiblingTo, Serena\_Williams) & \\
\cdashline{2-4}
 & $T_i$: (Person A, IsMarriedTo, Person B) & (Hillary\_Clinton, IsMarriedTo, Bill\_Clinton) & \multirow{2}{*}{352} \\
 & $T_j$: (Person B, IsMarriedTo, Person A) & (Bill\_Clinton, IsMarriedTo, Hillary\_Clinton) & \\
\cdashline{2-4}
 & $T_i$: (Person A, HasAFriendshipWith, Person B) & (Bob\_Dylan, HasAFriendshipWith, The\_Beatles) & \multirow{2}{*}{1,832} \\
 & $T_j$: (Person B, HasAFriendshipWith, Person A) & (The\_Beatles, HasAFriendshipWith, Bob\_Dylan) & \\
\cdashline{2-4}
 & $T_i$: (Person A, IsAPeerOf, Person B) & (Jimi\_Hendrix, IsAPeerOf, Eric\_Clapton) & \multirow{2}{*}{132} \\
 & $T_j$: (Person B, IsAPeerOf, Person A) & (Eric\_Clapton, IsAPeerOf, Jimi\_Hendrix) & \\
\cdashline{1-4}
\multirow{6}{*}{$\langle T_i$, ImpliesLocation, $T_j \rangle$} & $T_i$: (Location A, Contains, Location B) & (England, Contains, Warwickshire) & \multirow{2}{*}{2,415} \\
 & $T_j$: (Location containing A, Contains, Location in B) & (United\_Kingdom, Contains, Birmingham) & \\
\cdashline{2-4}
 & $T_i$: (Organization A, Headquarter, Location B) & (Kyoto\_University, Headquarter, Kyoto) & \multirow{2}{*}{820} \\
 & $T_j$: (Location B, Contains, Organization A) & (Kyoto, Contains, Kyoto\_University) & \\
\cdashline{2-4}
 & $T_i$: (Country A, CapitalIsLocatedIn, City B) & (Sweden, CapitalIsLocatedIn, Stockholm) & \multirow{2}{*}{83} \\
 & $T_j$: (Country A, Contains, City B) & (Sweden, Contains, Stockholm) & \\
\cdashline{1-4}
\multirow{10}{*}{$\langle T_i$, ImpliesProfession, $T_j \rangle$} & $T_i$: (Person A, IsA, Specialized Profession of B) & (Mariah\_Carey, IsA, Singer-songwriter) & \multirow{2}{*}{2,364} \\
 & $T_j$: (Person A, IsA, Profession B) & (Mariah\_Carey, IsA, Musician) & \\
\cdashline{2-4}
 & $T_i$: (Rock\&Roll Hall of Fame, Inducts, Person A) & (Rock\&Roll\_Hall\_of\_Fame, Inducts, Bob\_Dylan) & \multirow{2}{*}{66} \\
 & $T_j$: (Person A, IsA, Musician/Artist) & (Bob\_Dylan, IsA, Musician) & \\
\cdashline{2-4}
 & $T_i$: (Film A, IsWrittenBy, Person B) & (127\_Hours, IsWrittenBy, Danny\_Boyle) & \multirow{2}{*}{893} \\
 & $T_j$: (Person B, IsA, Writer/Film Producer) & (Danny\_Boyle, IsA, Film\_producer) & \\
\cdashline{2-4}
 & $T_i$: (Person A, ActsIn, Film B) & (Liam\_Neeson, ActsIn, Love\_Actually) & \multirow{2}{*}{10,864} \\
 & $T_j$: (Person A, IsA, Actor) & (Liam\_Neeson, IsA, Actor) & \\
\cdashline{2-4}
 & $T_i$: (Person A, HoldsPosition, Government Position B) & (Barack\_Obama, HoldsPosition, President) & \multirow{2}{*}{120} \\
 & $T_j$: (Person A, IsA, Politician) & (Barack\_Obama, IsA, Politician) & \\
\cdashline{1-4}
\multirow{6}{*}{$\langle T_i$, ImpliesSports, $T_j \rangle$} & $T_i$: (Team A, HasPosition, Position of B) & (Boston\_Red\_Socks, HasPosition, Infield) & \multirow{2}{*}{2,936} \\
 & $T_j$: (Team A, Plays, Sports B) & (Boston\_Red\_Socks, Plays, Baseball) & \\
\cdashline{2-4}
 & $T_i$: (League of A, Includes, Team B) & (National\_League, Includes, New\_York\_Mets) & \multirow{2}{*}{824} \\
 & $T_j$: (Team B, Plays, Sports A) & (New\_York\_Mets, Plays, Baseball) & \\
\cdashline{2-4}
 & $T_i$: (Team A, ParticipatesIn, Draft of B) & (Atlanta\_Braves, ParticipatesIn, MLB\_Draft) & \multirow{2}{*}{528} \\
 & $T_j$: (Team A, Plays, Sports B) & (Atlanta\_Braves, Plays, Baseball) & \\
\cdashline{1-4}
\multirow{2}{*}{$\langle T_i$, NextEventPlace, $T_j \rangle$} & $T_i$: (Event A, IsHeldIn, Location A) & (1932\_Summer\_Olympics, IsHeldIn, Los\_Angeles) & \multirow{2}{*}{47} \\
 & $T_j$: (Next Event of A, IsHeldIn, Location B) & (1936\_Summer\_Olympics, IsHeldIn, Berlin) & \\
\cdashline{1-4}
\multirow{2}{*}{$\langle T_i$, WorksFor, $T_j \rangle$} & $T_i$: (Person A, HoldsPosition, Vice President) & (Joe\_Biden, HoldsPosition, Vice\_President) & \multirow{2}{*}{13} \\
 & $T_j$: (Person B, HoldsPosition, President) & (Barack\_Obama, HoldsPosition, President) & \\
\cdashline{1-4}
\multirow{2}{*}{$\langle T_i$, SucceededBy, $T_j \rangle$} & $T_i$: (Person A, HoldsPosition, President/Vice President) & (George\_W.\_Bush, HoldsPosition, President) & \multirow{2}{*}{30} \\
 & $T_j$: (Person B, HoldsPosition, President/Vice President) & (Barack\_Obama, HoldsPosition, President) & \\
\cdashline{1-4}
\multirow{2}{*}{$\langle T_i$, TransfersTo, $T_j \rangle$} & $T_i$: (Person A, PlaysFor, Team B) & (David\_Beckham, PlaysFor, Real\_Madrid\_CF) & \multirow{2}{*}{377} \\
 & $T_j$: (Person A, PlaysFor, Team C) & (David\_Beckham, PlaysFor, LA\_Galaxy) & \\
\cdashline{1-4}
\multirow{2}{*}{$\langle T_i$, HigherThan, $T_j \rangle$} & $T_i$: (Item A, IsRankedIn, Ranking List C) & (Walmart, IsRankedIn, Fortune\_500) & \multirow{2}{*}{7,459} \\
 & $T_j$: (Item B, IsRankedIn, Ranking List C) & (Bank\_of\_America, IsRankedIn, Fortune\_500) & \\
\bottomrule
\end{tabular}
\caption{All types of nested triples to create \fbh and \fbhe. Table taken from \cite{DBLP:conf/aaai/Chanyoung}.}
\label{tb:fbh_full}
\end{table*}

\begin{table*}[htbp]
\scriptsize
\centering
\setlength{\tabcolsep}{0.45em}
\begin{tabular}{cllc}
\toprule
 & & Example & Frequency \\
\midrule
\multirow{6}{*}{$\langle T_i$, EquivalentTo, $T_j \rangle$} & $T_i$: (Person A, IsMarriedTo, Person B) & (Hillary\_Clinton, IsMarriedTo, Bill\_Clinton) & \multirow{2}{*}{314} \\
 & $T_j$: (Person B, IsMarriedTo, Person A) & (Bill\_Clinton, IsMarriedTo, Hillary\_Clinton) & \\
\cdashline{2-4}
 & $T_i$: (Location A, UsesLanguage, Language B) & (Brazil, UsesLanguage, Portuguese\_Language) & \multirow{2}{*}{120} \\
 & $T_j$: (Language B, IsSpokenIn, Location A) & (Portuguese\_Language, IsSpokenIn, Brazil) & \\
\cdashline{2-4}
 & $T_i$: (Person A, Influences, Person B) & (Baruch\_Spinoza, Influences, Immanuel\_Kant) & \multirow{2}{*}{394} \\
 & $T_j$: (Person B, IsInfluencedBy, Person A) & (Immanuel\_Kant, IsInfluencedBy, Baruch\_Spinoza) & \\
\cdashline{1-4}
\multirow{4}{*}{$\langle T_i$, ImpliesLanguage, $T_j \rangle$} & $T_i$: (Location A, HasOfficialLanguage, Language B) & (Italy, HasOfficialLanguage, Italian\_Language) & \multirow{2}{*}{196} \\
 & $T_j$: (Location A, UsesLanguage, Language B) & (Italy, UsesLanguage, Italian\_Language) & \\
\cdashline{2-4}
 & $T_i$: (Location A, UsesLanguage, Language B) & (United\_States, UsesLanguage, English\_Language) & \multirow{2}{*}{75} \\
 & $T_j$: (Location in A, UsesLanguage, Language B) & (California, UsesLanguage, English\_Language) & \\
\cdashline{1-4}
\multirow{12}{*}{$\langle T_i$, ImpliesProfession, $T_j \rangle$} & $T_i$: (Work A, MusicComposedBy, Person B) & (Forrest\_Gump, MusicComposedBy, Alan\_Silvestri) & \multirow{2}{*}{553} \\
 & $T_j$: (Person B, IsA, Musician/Composer) & (Alan\_Silvestri, IsA, Composer) & \\
\cdashline{2-4}
 & $T_i$: (Work A, Starring, Person B) & (Love\_Actually, Starring, Liam\_Neeson) & \multirow{2}{*}{737} \\
 & $T_j$: (Person B, IsA, Actor) & (Liam\_Neeson, IsA, Actor) & \\
\cdashline{2-4}
 & $T_i$: (Work A, CinematographyBy, Person B) & (Jurassic\_Park, CinematographyBy, Dean\_Cundey) & \multirow{2}{*}{299} \\
 & $T_j$: (Person B, IsA, Cinematographer) & (Dean\_Cundey, IsA, Cinematographer) & \\
\cdashline{2-4}
 & $T_i$: (Work A, IsDirectedBy, Person B) & (Psycho, IsDirectedBy, Alfred\_Hitchcock) & \multirow{2}{*}{295} \\
 & $T_j$: (Person B, IsA, Film\_Director/Television\_Director) & (Alfred\_Hitchcock, IsA, Film\_Director) & \\
\cdashline{2-4}
 & $T_i$: (Work A, IsProducedBy, Person B) & (King\_Kong, IsProducedBy, Merian\_C.\_Cooper) & \multirow{2}{*}{354} \\
 & $T_j$: (Person B, IsA, Film\_Producer/Television\_Producer) & (Merian\_C.\_Cooper, IsA, Film\_Producer) & \\
\cdashline{2-4}
 & $T_i$: (Person A, AssociatesWithRecordLabel, Record B) & (Bo\_Diddley, AssociatesWithRecordLabel, Atlantic\_Records) & \multirow{2}{*}{155} \\
 & $T_j$: (Person A, IsA, Record\_Producer) & (Bo\_Diddley, IsA, Record\_Producer) & \\
\cdashline{1-4}
\multirow{6}{*}{$\langle T_i$, ImpliesLocation, $T_j \rangle$} & $T_i$: (Location A, IsPartOf, Location B) & (Ann\_Arbor, IsPartOf, Washtenaw\_County\_Michigan) & \multirow{2}{*}{1,174} \\
 & $T_j$: (Location A, IsPartOf, Location Containing B) & (Ann\_Arbor, IsPartOf, Michigan) & \\
\cdashline{2-4}
 & $T_i$: (Organization A, IsLocatedIn, Location B) & (Adobe\_Systems, IsLocatedIn, San\_Jose\_California) & \multirow{2}{*}{250} \\
 & $T_j$: (Organization A, IsLocatedIn, Location Containing B) & (Adobe\_Systems, IsLocatedIn, California) & \\
\cdashline{2-4}
 & $T_i$: (Person A, LivesIn, Location B) & (Mariah\_Carey, LivesIn, New\_York\_City) & \multirow{2}{*}{213} \\
 & $T_j$: (Person A, LivesIn, Location Containing B) & (Mariah\_Carey, LivesIn, New\_York) & \\
\cdashline{1-4}
\multirow{2}{*}{$\langle T_i$, ImpliesTimeZone, $T_j \rangle$} & $T_i$: (Location A, TimeZone, Time Zone B) & (Czech\_Republic, TimeZone, Central\_European\_Time) & \multirow{2}{*}{409} \\
 & $T_j$: (Location in A, TimeZone, Time Zone B) & (Prague, TimeZone, Central\_European\_Time) & \\
\cdashline{1-4}
\multirow{2}{*}{$\langle T_i$, ImpliesGenre, $T_j \rangle$} & $T_i$: (Musician A, Genre, Genre B) & (Pharrell\_Williams, Genre, Progressive\_Rock) & \multirow{2}{*}{767} \\
 & $T_j$: (Musician A, Genre, Parent Genre of B) & (Pharrell\_Williams, Genre, Rock\_Music) & \\
\cdashline{1-4}
\multirow{2}{*}{$\langle T_i$, NextAlmaMater, $T_j \rangle$} & $T_i$: (Person A, StudiesIn, Institution B) & (Gerald\_Ford, StudiesIn, University\_of\_Michigan) & \multirow{2}{*}{112} \\
 & $T_j$: (Person A, StudiesIn, Institution C) & (Gerald\_Ford, StudiesIn, Yale\_University) & \\
\cdashline{1-4}
\multirow{2}{*}{$\langle T_i$, TransfersTo, $T_j \rangle$} & $T_i$: (Person A, PlaysFor, Team B) & (Ronaldo, PlaysFor, FC\_Barcelona) & \multirow{2}{*}{300} \\
 & $T_j$: (Person A, PlaysFor, Team C) & (Ronaldo, PlaysFor, Inter\_Millan) & \\
\bottomrule
\end{tabular}
\caption{All types of nested triples to create \dbhe. Table taken from \cite{DBLP:conf/aaai/Chanyoung}.}
\label{tb:dbh_full}
\end{table*}

\end{document}